\documentclass[journal]{IEEEtran}

\ifCLASSINFOpdf
\else
\fi

\usepackage{cite}
\usepackage[capitalize]{cleveref}
\usepackage{times}
\usepackage{epsfig}
\usepackage{graphicx}
\usepackage[cmex10]{amsmath}
\usepackage{amssymb}
\usepackage{color}
\usepackage{xcolor}
\usepackage{multirow}
\usepackage{array}
\usepackage{epstopdf}
\usepackage{subfig}
\usepackage{caption}
\usepackage[font=footnotesize]{caption}
\usepackage{booktabs}
\usepackage{enumerate}
\usepackage{float}

\newcommand{\etal}{\textit{et al}. }

\usepackage{eso-pic}

\begin{document}
%

\AddToShipoutPictureBG*{%
    \AtPageUpperLeft{%
        \setlength\unitlength{1in}%
        \hspace*{\dimexpr0.5\paperwidth\relax}
        \makebox(0,-0.25)[c]{\small This work has been submitted to the IEEE for possible publication.}
        \makebox(0,-0.5)[c]{\small Copyright may be transferred without notice, after which this version may no longer be accessible.}%
    }}

\title{Single Image Super-Resolution via Cascaded Multi-Scale Cross Network}

\author{Yanting~Hu,  
             Xinbo~Gao,~\IEEEmembership{Senior Member,~IEEE,}
             Jie~Li,
             Yuanfei~Huang,
             and ~Hanzi Wang,~\IEEEmembership{Senior Member,~IEEE}
\thanks{Y. Hu, J. Li and Y. Huang are with the Video and Image Processing System Laboratory, School of Electronic Engineering, Xidian University, Xi¡¯an 710071, China (email: {yantinghu2012@gmail.com}; {leejie@mail.xidian.edu.cn}; \href{mailto:yf_huang@stu.xidian.edu.cn}{yf\_huang@stu.xidian.edu.cn}).}
\thanks{X. Gao is with the State Key Laboratory of Integrated Services Networks, School of Electronic Engineering, Xidian University, Xi¡¯an 710071, China (e-mail: {xbgao@mail.xidian.edu.cn}).}
\thanks{H. Wang is with the Fujian Key Laboratory of Sensing and Computing for Smart City, School of Information Science and Engineering,  Xiamen University, China (e-mail: {wang.hanzi@gmail.com}).}}
\maketitle

\begin{abstract}
  The deep convolutional neural networks have achieved significant improvements in accuracy and speed for single image super-resolution. However, as the depth of network grows, the information flow  is weakened and the training becomes harder and harder. On the other hand, most of the models adopt a single-stream structure with which integrating complementary contextual information under different receptive fields is difficult. To improve information flow and to capture sufficient knowledge for reconstructing the high-frequency details, we propose a cascaded multi-scale cross network (CMSC) in which a sequence of subnetworks is cascaded to infer high resolution features in a coarse-to-fine manner. In each cascaded subnetwork, we stack multiple multi-scale cross (MSC) modules to fuse complementary multi-scale information in an efficient way as well as to improve information flow across the layers. Meanwhile, by introducing residual-features learning in each stage, the relative information between high-resolution  and low-resolution features is fully utilized to further boost reconstruction performance. We train the proposed network with cascaded-supervision and then assemble the intermediate predictions of the cascade to achieve high quality image reconstruction. Extensive quantitative and qualitative evaluations on benchmark datasets illustrate the superiority of our proposed method over  state-of-the-art super-resolution methods.
\end{abstract}


\IEEEpeerreviewmaketitle

\section{Introduction}

\IEEEPARstart{S}{ingle} image super-resolution (SR) aims to infer a high resolution (HR) image from one low resolution (LR) input image, which has a wide applications in video surveillance, remote sensing imaging, medical imaging and digital entertainment. Since the SR process is inherently an ill-posed inverse problem, exploring and enforcing strong prior information about the HR image are necessary to guarantee the stability of the SR process. Many traditional example-based SR methods have been devoted to resolving this problem via probabilistic graphical model  \cite{1Freeman2000IJCV,  2Polatkan2015TPAMI}, neighbor embedding \cite{3Chang2004CVPR, 4Gao2012TIP}, sparse coding \cite{5Yang2010TIP,  6He2013CVPR}, linear or nonlinear regression \cite{7Timofte2014ACCV,  8Hu2016TIP,  9Wang2016TIP} and random forest \cite{10Schulter2015CVPR}.

\begin{figure}[t]
\vspace{-0.2cm}
\captionsetup{belowskip=-14pt}
\includegraphics[width=1.0\linewidth]{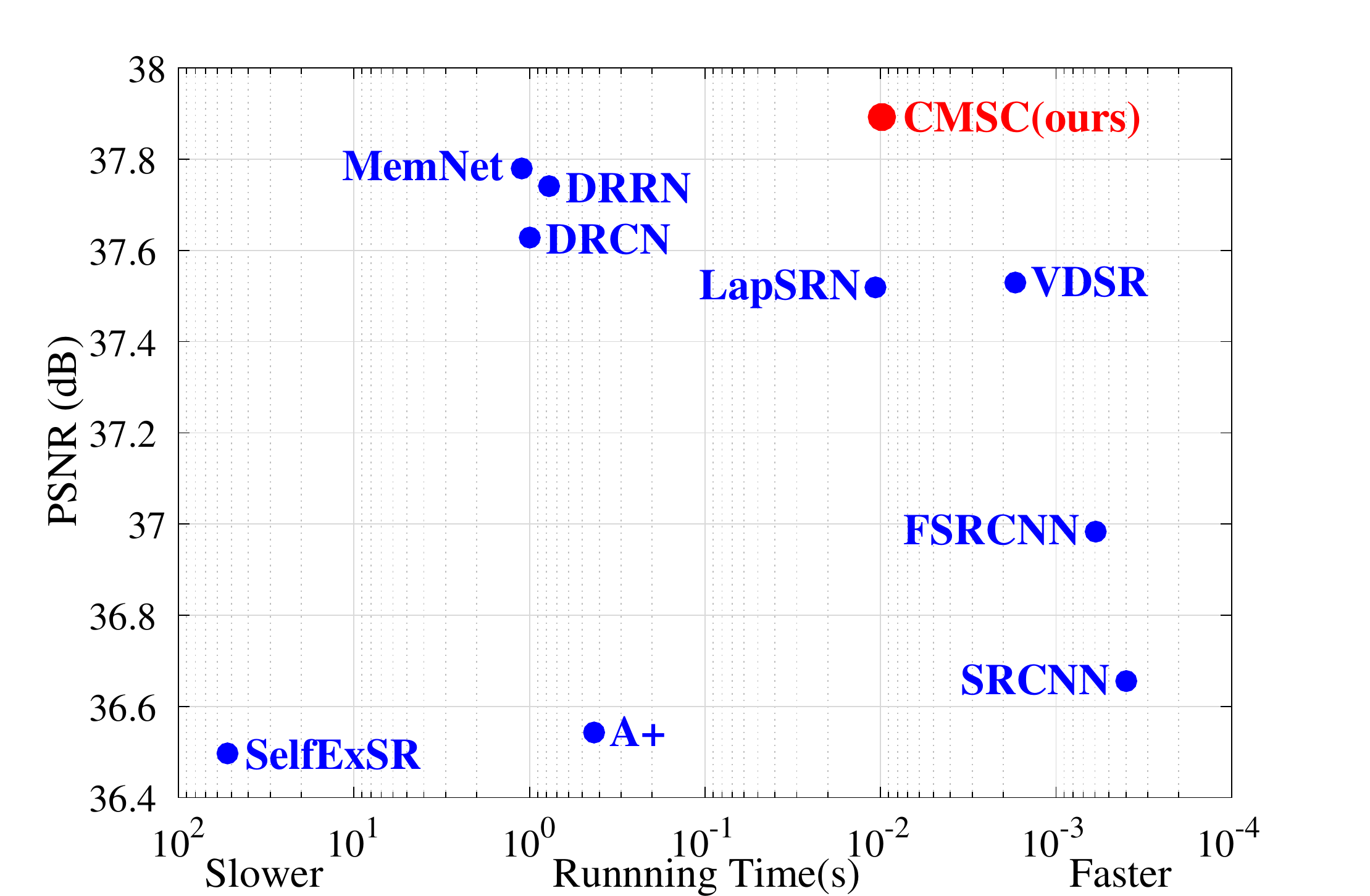}
\caption{PSNR performance versus runtime (evaluated in seconds). The results are evaluated on  the Set5 dataset  for a scale factor of $2\times$. The proposed CMSC achieves the best performance with relatively less execution time.}
\label{fig:PSNR performance versus runtime}
\vspace{-0.2cm}
\end{figure}

\begin{figure*}
\vspace{-0.2cm}
\captionsetup[subfigure]{farskip = 0pt}
\captionsetup{belowskip=-16pt}
    \small
    \begin{minipage}[b]{1\linewidth}
        \centering
        \subfloat[The architecture of the proposed CMSC network]{
            \centering
            \includegraphics[width=1\linewidth]{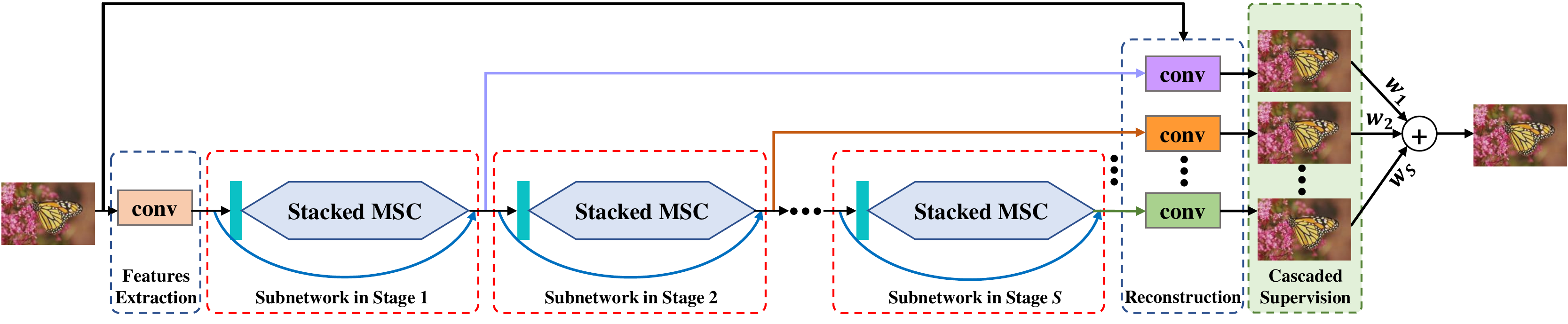}
        }\\
        \vspace{0.2cm}
        \subfloat[The architecture of each cascaded subnetwork in CMSC network]{
            \centering
            \includegraphics[width=1\linewidth]{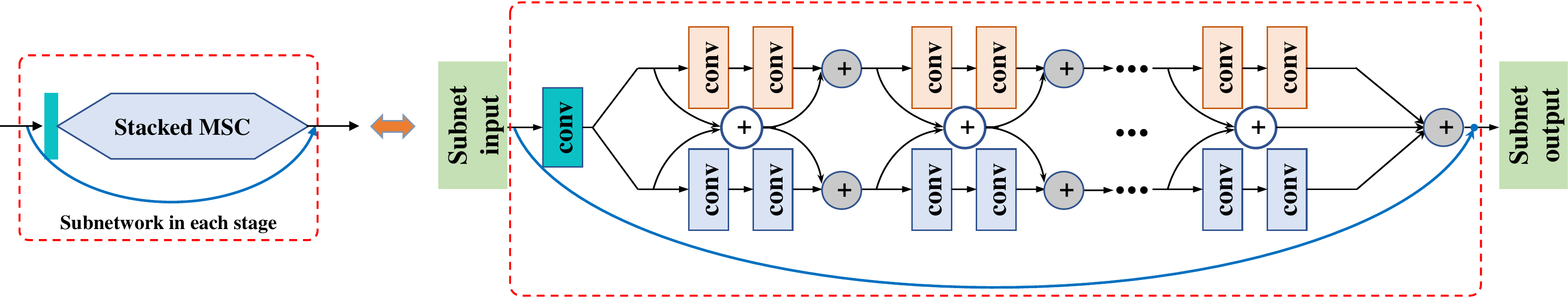}
        }
        \end{minipage}
\caption{ The architectures of our proposed CMSC model and each cascaded subnetwork in CMSC model. (a) The overall architecture of the proposed CMSC model, which utilizes cascaded structure, cascaded-supervision and predictions assembling to boost SR performance. (b) The architecture of each subnetwork in (a), which applies a sequence of stacked MSC modules to capture sufficient information for inferring the HR features. The {\color{blue}blue} arrow denotes skip connection for residual-features learning.}
\label{fig:The architectures}
\end{figure*}

More recently, deep networks have been utilized for image SR via modeling the mapping from LR to HR space and achieved impressive results. Dong \etal \cite{11Dong2016TPAMI}  present a deep convolutional neural network (CNN) with three convolutional  layers (SRCNN) to predict the nonlinear relationship between LR  and HR images. Due to the slow convergence and the difficulty in deeper network training, their deeper networks with more convolutional layers do not perform better. To break through the limitation of SRCNN, Kim \etal  \cite{12KimJ2016CVPR_VDSR}  propose a very deep convolutional network (VDSR) for highly accurate SR and adopt extremely high learning rate as well as residual-learning to speed-up the training process. Besides, they use adjustable gradient clipping to solve gradient explosion problem. Meanwhile, to control the number of model parameters in deeper network, Kim \etal  \cite{13KimJ2016CVPR_DRCN}  also propose a deeply-recursive convolutional network (DRCN) in which recursive-supervision and skip-connection are used to ease the difficulty of training. For the same reason, the deep recursive residual network (DRRN) is proposed by Tai \etal \cite{14TaiY2017CVPR_DRRN}, in which  global and local residual learning as well as recursive module are introduced to reduce the number of  model parameters. Since the identity mapping in residual network (ResNet) \cite{15HeK2016ECCV_identity}  makes the training of very deep networks easy, ResNet architecture with identity mapping has been applied in image restoration. Ledig \etal \cite{16LedigC2017CVPR_SRResNet}  employ a deep residual network with 16 residual blocks (SRResNet) and skip-connection to super-resolve LR image with an upscaling factor of $4\times$. Lim \etal  \cite{17Lim2017CVPRW_EDSR} develop an enhanced deep super-resolution network by removing the batch normalization layers in SRResNet and their method win the NTIRE2017 super-resolution challenge \cite{18Timofte2017CVPRW_NTIRE}. Further, to conveniently pass information across several layers or modules, the pattern of multiple or dense skip connections between layers or modules is adopted in SR. Mao \etal \cite{19Mao2016NIPS_RED}  propose a 30-layer convolutional residual encoder-decoder network (RED30) for image restoration, which uses skip-layer connection to symmetrically link convolutional layers and deconvolutional layers. Inspired by densely connected convolutional networks (DenseNet) \cite{20Huang2017CVPR_DenseNet} which achieves high performance in image classification, Tong \etal \cite{21Tong2017ICCV_DenseSR} utilize the DenseNet structure as building blocks to reuse learnt feature maps and introduce dense skip connections to fuse features at different levels. Meanwhile, Tai \etal  \cite{22TaiY2017ICCV_MemNet}  propose the deepest persistent memory network (MemNet) for image restoration, in which a memory block is applied to achieve persistent memory and multiple memory blocks are stacked with a densely connected structure to ensure maximum information flow between blocks.

Assembling a set of independent subnetworks, widely adopted in image SR, is also an effective solution to improve SR performance. Wang \etal  \cite{23WangY2016arXiv}  explore a new deep CNN architecture by jointly training both deep and shallow CNNs, where the shallow network stabilizes training and the deep network ensures accurate HR reconstruction. Similarly, Tang \etal \cite{24Tang2018Neurocomputing}  propose a joint residual network with three parallel subnetworks to learn the low frequency information and high frequency information for SR. Yamanaka \etal \cite{25Yamanaka2017ICONIP}  combine skip connection layers and parallelized CNNs into a deep CNN architecture for image reconstruction. Also,  \cite{26Ren2017CVPRW}  discusses different subnetworks fusion schemes and proposes a context-wise network fusion approach which integrates the outputs of individual networks by additional convolutional layers. In addition to parallel network fusion, progressive network fusion or cascaded networks structure is adopted in SR. Wang \etal \cite{27Wang2015ICCV}  exploit the natural sparsity of images and build a cascaded sparse coding network in which each subnetwork is trained for a small upscaling factor. Recently, a cascade of convolutional neural networks is also utilized in \cite{28Cui2014ICCV}  and \cite{29LaiWS2017CVPR_LapSRN}  to progressively upscale low-resolution images.

For image SR, the input and output images as well as information among layers in networks are highly correlated. It is important for SR to combine the knowledge of features at different levels and different scales. Although the previous SR approaches utilize deeper networks to take more contextual information, the fusion of complementary multi-scale information under different receptive fields is still difficult due to adopting single-stream structure in their designed networks, such as SRCNN  \cite{11Dong2016TPAMI}, FSRCNN  \cite{46Dong2016ECCVFSRCNN}, VDSR \cite{12KimJ2016CVPR_VDSR}  and DRCN  \cite{13KimJ2016CVPR_DRCN}. Besides, it is found in \cite{30Zagoruyko2017arXivDiracNet,  31Zhao2017arXivMR}  that for deep networks, as a way to increase the number of layers, increasing the width is more effective than increasing the depth. Therefore, in order to conveniently promote information integration for image SR, the multi-stream structure and network widening may be beneficial. On the other hand, applying the residual information learning and the cascaded or progressive structure into SR can simplify the difficulty of direct super-resolving images, which has been manifested in several SR methods. Taking the above into consideration, we propose a deep Cascaded Multi-Scale Cross  network (CMSC) for SR (illustrated in Fig.\ref{fig:The architectures}), which consists of a feature extraction network, a set of cascaded subnetworks and a reconstruction network. A set of cascaded subnetworks is utilized to gradually reconstruct HR features. In each stage of the cascade, we develop a multi-scale cross (MSC) module with two branches (depicted in Fig.\ref{fig:The built modules}(b)) for fusing multi-level information under different receptive fields, and then stack the MSC modules as a subnetwork (shown in Fig.\ref{fig:The architectures}(b)) for learning the residual information between HR and LR features. During training, the multiple cascaded-supervised strategy is adopted to supervise all of the predictions from subnetworks and the final output from overall CMSC model. Compared with  state-of-the-art SR methods, our proposed CMSC network achieves the best performance with relatively less execution time, as illustrated in Fig.\ref{fig:PSNR performance versus runtime}. In summary, the major contributions of our proposed method include:

1) A multi-scale cross module not only to fuse multi-scale complementary information under different receptive fields but also to help information flow across the network. In the multi-scale cross module, two branches having different receptive fields are assembled in parallel via averaging and adding.

2) A subnetwork with residual-features learning to reconstruct the high-resolution features. In the subnetwork, multiple multi-scale cross modules are stacked in sequence and an identity mapping is used to add the input to the end of the subnetwork. Therefore, instead of inferring the direct mapping from LR  to HR features, the subnetwork uses the stacked multi-scale cross modules to reconstruct the residual features.

3) A cascaded networks structure to gradually decrease the gap between estimated HR features and ground truth HR features. Several residual-features learning subnetworks are cascaded to reconstruct the HR features with a coarse-to-fine manner. All of the outputs from the cascaded stages are fed into the corresponding reconstruction networks to obtain the intermediate predictions, which are utilized to compute the final prediction via weighted average. Both the intermediate predictions and final prediction are supervised in training. Thus,
the SR performance is boosted by the cascaded-supervision and the assembling of intermediate predictions.

The remainder of this paper is organized as follows. Section II discusses the related  single image SR methods  and  network architectures. Section III describes the proposed CMSC network for SR in detail. Model analysis and experimental comparison with other state-of-the-art methods are presented in Section IV, and Section V concludes the paper with observations and discussion.

\section{Related Work}

Numerous single image SR methods and different network architectures have been proposed in the literatures. Here, we focus our discussions on the approaches which are relative to our method.

\subsection{Multi-branch Module}
\begin{figure}[t]
\captionsetup[subfigure]{farskip = 0pt}
\captionsetup{belowskip=-16pt}
    \small
    \begin{minipage}[b]{1\linewidth}
        \centering
        \subfloat[MR module]{
            \centering
            \includegraphics[width=0.3\linewidth]{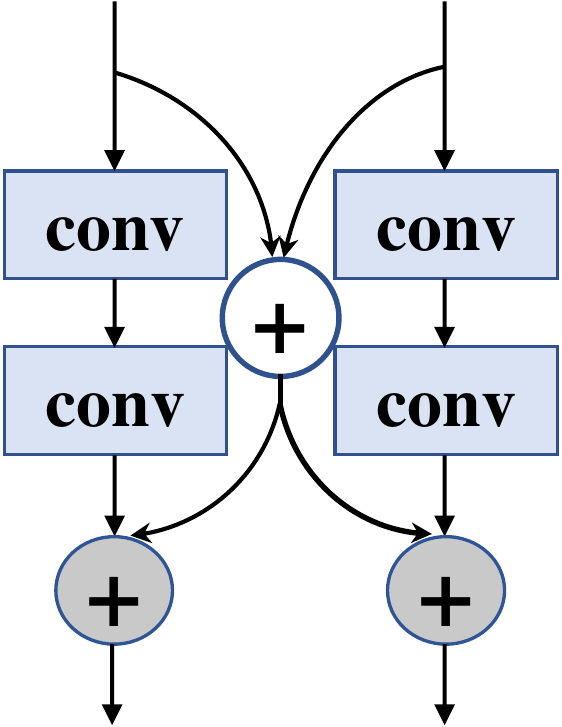}
        }
        \hspace{0.1\linewidth}
        \subfloat[MSC module]{
            \centering
            \includegraphics[width=0.3\linewidth]{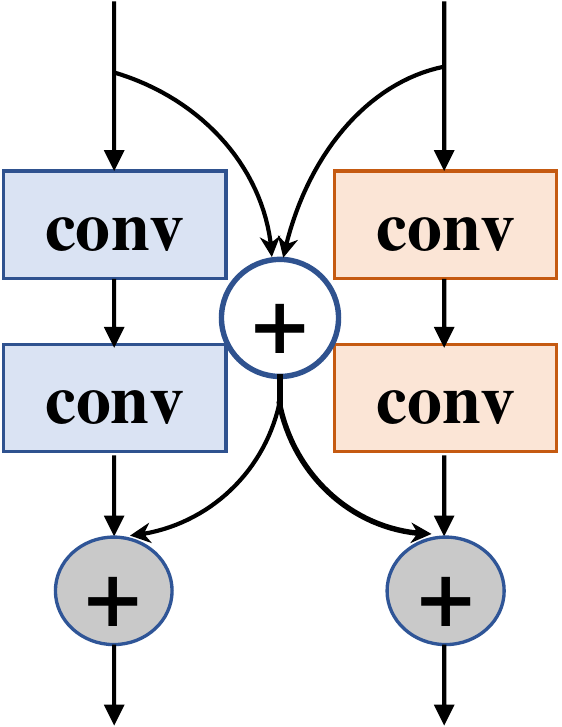}
        }
        \end{minipage}
\caption{The built modules: (a) A merge-and-run (MR) module, in which all convolutional filters have the same size. (b) A multi-scale cross (MSC) module, where the convolutional filters in two branches have different sizes and are denoted by different color rectangles for discrimination. The MSC module is designed to fuse information under different receptive fields. In (a) and (b), the ``$\text{+}$''  operators  in white circles denote the average operation and in gray circles denote the element-wise addition.}
\label{fig:The built modules}
\end{figure}

To improve information flow and to make training easy, the networks consisting of multi-branch modules have been developed, such as highway networks \cite{32Srivastava2015NIPS}, residual networks \cite{15HeK2016ECCV_identity,  33He2016CVPRResNet}, and GoogLeNet \cite{34Szegedy2015CVPR,  35Szegedy2016arXiv, 36Szegedy2016CVPR}. Residual network is built by stacking a sequence of residual blocks which contain a residual branch and an identity mapping. The inception module in GoogLeNet consists of parallel convolutional branches with different kernel sizes which are then concatenated for width increase and information fusion. More recently, Zhao \etal  \cite{31Zhao2017arXivMR}   propose a deep merge-and-run neural network, which contains a set of merge-and-run (MR) blocks (depicted in Fig.\ref{fig:The built modules}(a)). The MR block assembles residual branches in parallel with a merge-and-run mapping, which is shown as a linear idempotent function in \cite{31Zhao2017arXivMR}. Idempotent mapping implies that  the information from the early blocks can quickly flow to the later blocks and the gradient also can be quickly back-propagated to the early blocks from the later blocks, and thus the training difficulty is reduced. Although these different methods vary in network topology and training procedure, they all share a key characteristic: they contain multiple branches in each block. On the other hand, multi-branch networks can be viewed as an ensemble of many networks with different depths by which the performance can be boosted. Considering the advantages of  multi-branch module, we extend merge-and-run block via operating the convolutions with different kernel sizes on two parallel branches to fuse information at different scales for SR.

\vspace{-0.2cm}
\subsection{Residual Learning}
Since the SR prediction is largely similar to the input, residual learning is widely adopted in SR to achieve faster convergence and better accuracy. In \cite{8Hu2016TIP}, residual feature patches between the estimated HR feature patches and the ground truth feature patches are estimated via a set of cascaded linear regressions. In deep learning based method, two networks of VDSR \cite{12KimJ2016CVPR_VDSR} and DRCN \cite{13KimJ2016CVPR_DRCN}  are built to learn the residual image between HR and LR images by using a skip-connection from the input to the reconstruction layer. Later, the residual learning is extensively utilized in different SR networks \cite{21Tong2017ICCV_DenseSR,  22TaiY2017ICCV_MemNet, 23WangY2016arXiv, 24Tang2018Neurocomputing, 25Yamanaka2017ICONIP}. In \cite{23WangY2016arXiv}, instead of using bicubic interpolation to obtain the coarse HR images, the shallow network of three convolutional layers is applied to coarsely estimate HR images and then the deep network is utilized to predict the residual images between the ground truth HR images and the coarse HR images. Tai \etal  \cite{14TaiY2017CVPR_DRRN}  term the way of estimating the residual images by the skip-connection as global residual learning (GRL) (like in VDSR \cite{12KimJ2016CVPR_VDSR}  and DRCN \cite{13KimJ2016CVPR_DRCN}) and introduce a multi-path mode local residual learning which is combined with GRL to boost SR performance. Taking the effectiveness of GRL in training deep networks into account, we also adopt GRL in our proposed method. Further, we introduce the residual learning to the feature space, termed as residual-features learning (RFL), which is performed in each stage of the cascaded process.

\subsection{Progressive Structure}
In image SR, reconstructing the high-frequency details becomes very challenging when the upscaling factor is large. To simplify the difficulty of direct super-resolving the details, some progressive or cascaded structures for SR are proposed. There are two fashions in cascaded structure for SR: stage-by-stage refining and stage-by-stage upscaling. In former manner, the output of the previous stage is taken as the input and the ground truth as target for each stage, where the input and output have the same size. Meanwhile, the cascade minimizes the prediction errors at each stage and thus the prediction is gradually close to the target. Hu \etal \cite{8Hu2016TIP}  develop a cascaded linear regressions framework to refine the predicted feature patches in a coarse-to-fine fashion and merge all predicted patches to generate an HR image. In stage-by-stage upscaling manner, one stage of the cascade is utilized to upscale LR image with a smaller scale factor, the output of which is further fed into the next stage until the desired image scale. With this manner, deep network cascade is proposed in \cite{28Cui2014ICCV}  by using a cascade of multiple stacked collaborative local auto-encoders to gradually upscale low-resolution images. More recently, Lai \etal  \cite{29LaiWS2017CVPR_LapSRN}  present a Laplacian pyramid super-resolution network (LapSRN) based on a cascade of convolutional neural networks, which can progressively predict the sub-band residuals of HR images at multiple pyramid levels. To supervise intermediate stages in LapSRN, different scales of HR images at the corresponding levels need be generated by downsampling the ground truth HR image. Compared with the refining mode in which the cascade explicitly minimizes the prediction errors at each stage, this incremental approach has a loose control over the errors \cite{37Timofte2016CVPRSevenWays}. Therefore, we utilize the cascaded structure with a stage-by-stage refining fashion to gradually refine the HR features.

\section{The Proposed CMSC Network}

Our proposed CMSC model for SR, outlined in Fig.\ref{fig:The architectures}, consists of a feature extraction network, a set of cascaded subnetworks and a reconstruction network. The feature extraction network is applied to represent the input image as the feature maps via a convolutional layer. A set of cascaded subnetworks is designed to reconstruct the HR features from LR features with a coarse-to-fine manner. The reconstructed HR feature maps are then fed into the reconstruction network to generate the HR images via the convolution operation. In this section, we describe the proposed model in detail, from the multi-scale cross module to the residual-features learning subnetwork and finally the overall cascaded network.

\subsection{Multi-scale Cross Module}
Due to the identity mappings to skip residual branches in ResNet \cite{15HeK2016ECCV_identity}, a deep residual network is easy to train. Similarly, to further reduce the training difficulty and to improve information flow, deep merge-and-run neural networks, built by stacking a sequence of merge-and-run (MR) blocks, is proposed in \cite{31Zhao2017arXivMR} for the standard image classification task. Depicted in Fig.\ref{fig:The built modules}(a), the MR block assembles the residual branches in parallel with a merge-and-run mapping: average the inputs of these residual branches and add the average to the output of each residual branch as the input of the subsequent residual branch respectively.

Inspired by MR block and considering that different receptive fields of convolutional networks can provide different contextual information which is very important for SR, we propose a multi-scale cross (MSC) module to incorporate different information under different sizes of receptive fields. As shown in Fig.\ref{fig:The built modules}(b), an MSC module similarly integrates two residual branches in parallel with a merge-and-run mapping but operates different kernel sizes of convolutions on two branches, where different convolutions are differentiated by different color rectangles in Fig.\ref{fig:The built modules}(b). Each residual branch in MSC module contains two convolutional layers and each convolutional layer is followed by batch normalization (BN) \cite{38Timofte2015ICMLBN} and LeakyReLU \cite{39Maas2013ICMLLeakReLU}.  Moreover, the ``$\text{+}$''  operators  in gray circles in Fig.\ref{fig:The built modules} are between BN and LeakyReLU and denote the addition operation, while the ``$\text{+}$'' operators  in white circles denote the average operation. Thus, with this design, these branches can provide complementary contextual information at multiple scales which is further combined by the merge-and-run mapping. By denoting ${{H}^{b1}}$ and ${{H}^{b2}}$ as the transition functions of   two residual branches respectively, the proposed module can be represented in matrix form as below.
\begin{equation}
\begin{aligned}
\label{Eq: eq 1}
\left[ \begin{matrix}
   \mathbf{x}_{o}^{b1}  \\[4pt]
   \mathbf{x}_{o}^{b2}  \\
\end{matrix} \right]={} & G \left( \left[ \begin{matrix}
   \mathbf{x}_{i}^{b1}  \\[4pt]
   \mathbf{x}_{i}^{b2}  \\
\end{matrix} \right] \right) \\[4pt]
={} & \left[ \begin{matrix}
   {{H}^{b1}}\left( \mathbf{x}_{i}^{b1} \right)  \\[4pt]
   {{H}^{b2}}\left( \mathbf{x}_{i}^{b2} \right)  \\
\end{matrix} \right] + \frac{1}{2} \left[ \begin{matrix}
   \mathbf{I} & \mathbf{I}  \\[4pt]
   \mathbf{I} & \mathbf{I}  \\
\end{matrix}  \right] \left[  \begin{matrix}
   \mathbf{x}_{i}^{b1}  \\[4pt]
   \mathbf{x}_{i}^{b2}  \\
\end{matrix} \right],
\end{aligned}
\end{equation}
where $\mathbf{x}_{i}^{b1}$ and $\mathbf{x}_{i}^{b2}$ ( $\mathbf{x}_{o}^{b1}$ and $\mathbf{x}_{o}^{b2}$ ) are the inputs (outputs) of two residual branches of the module and $\mathbf{I}$ is identity matrix. $\mathbf{P}=\frac{1}{2}\left[ \begin{matrix} \mathbf{I} & \mathbf{I}  \\ \mathbf{I} & \mathbf{I}  \\ \end{matrix} \right]$ is the transformation matrix of the merge-and-run mapping. With this multi-scale information fusing structure, the proposed module can exploit a wide variety of contextual information to infer missing high-frequency components. On the other hand, as analyzed in \cite{31Zhao2017arXivMR}, the transformation matrix $\mathbf{P}$ in merge-and-run mapping is idempotent, which is similarly possessed by the proposed MSC module. The property of idempotent not only helps rich information flow across the different modules, but also encourages gradient back-propagation during training. Experimental analysis is described in Section IV \emph{C}.

\begin{figure*}
\vspace{-0.3cm}
\captionsetup[subfigure]{farskip = 0pt}
\captionsetup{belowskip=-16pt}
    \small
    \begin{minipage}[b]{1\linewidth}
        \centering
        \subfloat[Four modules]{
            \centering
            \includegraphics[width=0.49\linewidth]{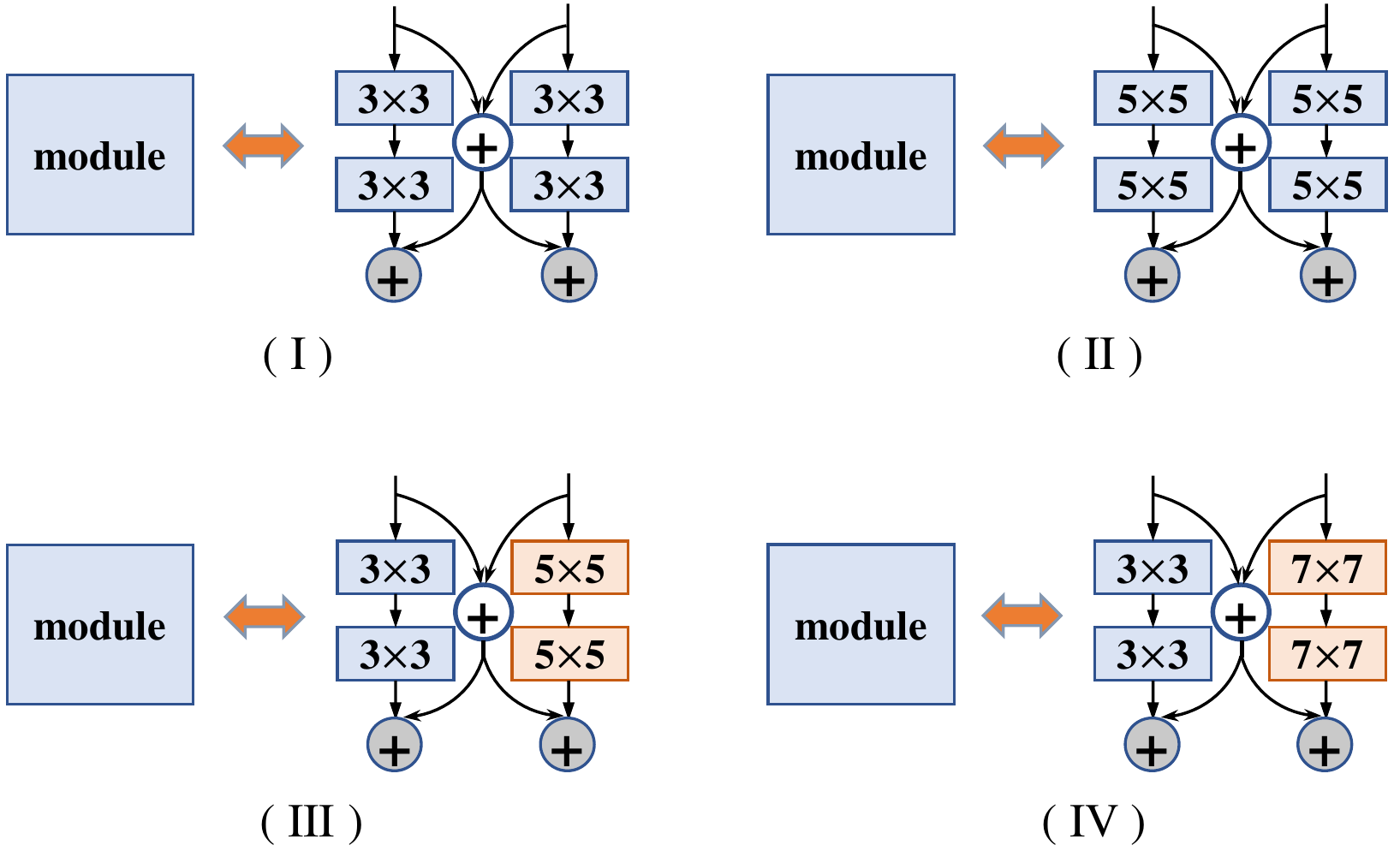}
        }
        \subfloat[Basic structure]{
            \includegraphics[height=0.4\linewidth]{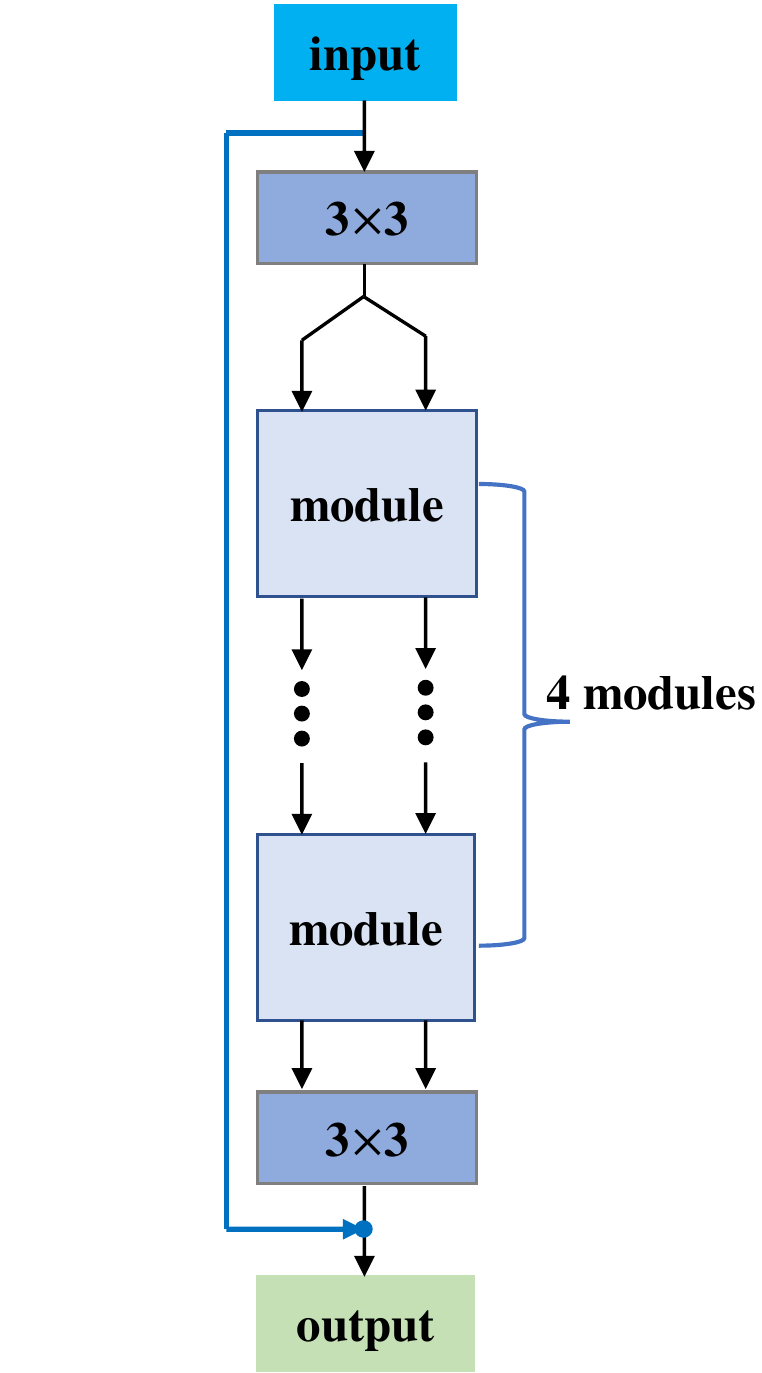}
        }
        \subfloat[VDSR]{
            \includegraphics[height=0.4\linewidth]{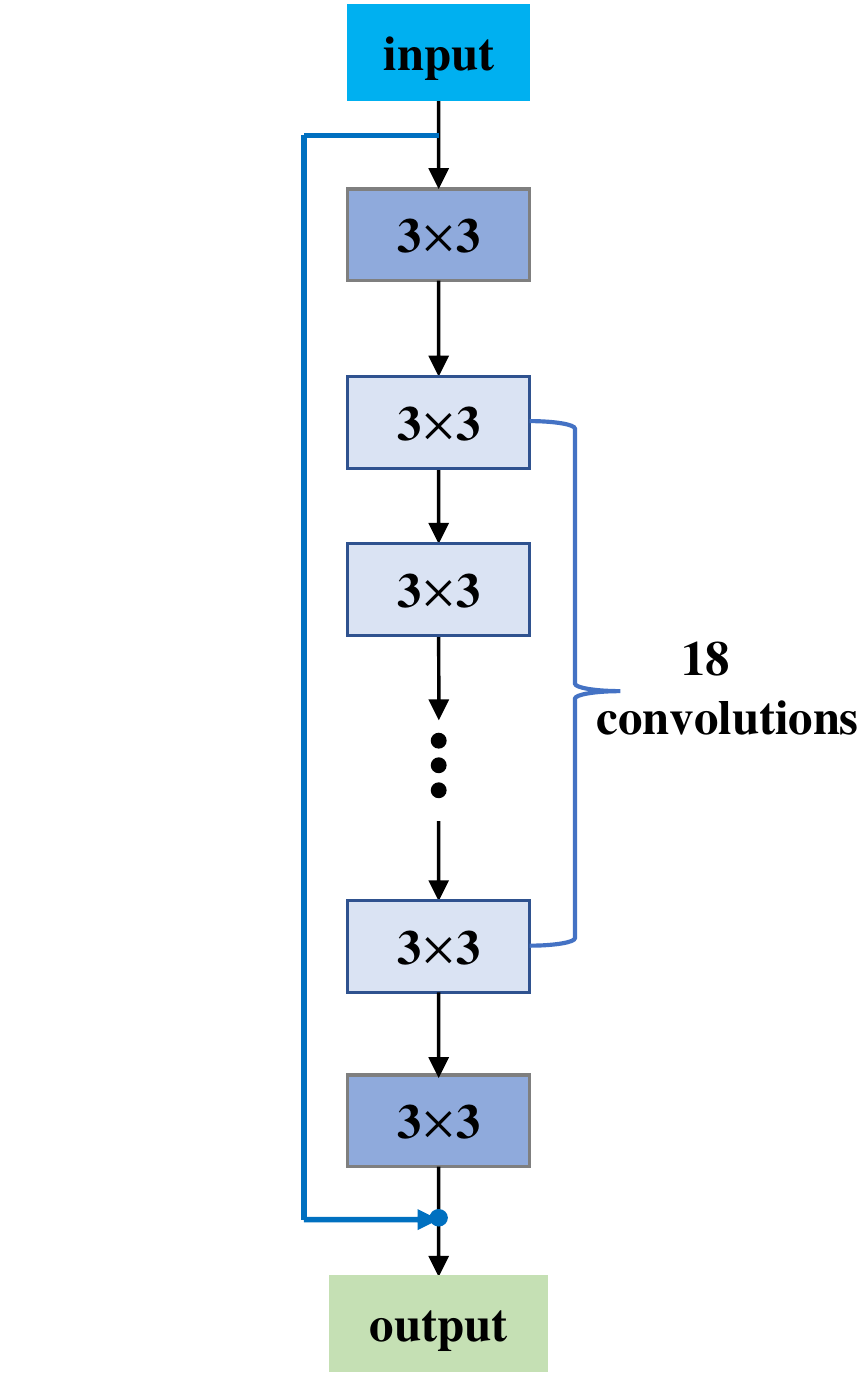}
        }
        \end{minipage}
\caption{Four network structures with different modules and the structure of VDSR. (a) Four built modules for comparison, including MR\_L3R3, MR\_L5R5, MSC\_L3R5 and MSC\_L3R7, which are depicted in (I), (II), (III) and (IV) respectively. (b) The built basic network structure with four modules, where all modules are replaced by one of the modules in (a) respectively to generate four corresponding networks for comparing SR performances. (c) The network structure of VDSR~\cite{12KimJ2016CVPR_VDSR}, which stacks 20 convolutional layers for image SR.}
\label{fig:Four network sructures}
\end{figure*}

\subsection{Residual-Features Learning Subnetwork}

Aiming to infer the HR features, the subnetwork with the residual-features learning (RFL) is built. As illustrated in Fig.\ref{fig:The architectures}(b), the building subnetwork contains a convolutional layer and a sequence of multi-scale cross (MSC) modules. At the end of the last MSC module, two outputs from two branches and the average of their inputs are fused by addition, which can be formulated as follows.
\vspace{-0cm}
\begin{equation}
\begin{aligned}
\label{Eq: eq 2}
{\mathbf{x}}_{o}={} & {G}^{last} \left( \left[ \begin{matrix}
   \mathbf{x}_{i}^{b1}  \\[4pt]
   \mathbf{x}_{i}^{b2}  \\
\end{matrix} \right] \right)\\
={} & {{H}^{b1}}\left( \mathbf{x}_{i}^{b1} \right)+{{H}^{b2}}\left( \mathbf{x}_{i}^{b2} \right)+\frac{1}{2}\left( \mathbf{x}_{i}^{b1}+\mathbf{x}_{i}^{b2} \right),
\end{aligned}
\end{equation}
where ${{\mathbf{x}}_{o}}$ denotes the output from the last MSC module and the other notations have the same meanings as those in Eq.~(\ref{Eq: eq 1}).  The first convolutional layer in the subnetwork is utilized to transform the input of the subnetwork as a specified number of feature maps. Then, with the parallel and intersecting structure, the stacked multi-scale cross (MSC) modules can be used to process these features via different processing procedures. Meanwhile, multiple merge-and-run mappings in subnetwork are exploited to integrate different information coming from two branches as well as to create quick paths directly sending the information of the intermediate branches to the later modules.

Since the features of HR image are highly correlated with the features of corresponding LR image, we introduce residual-features learning into our subnetwork by adding an identity branch from the input to the end of the subnetwork (blue  curves in Fig.\ref{fig:The architectures}(b)). Therefore, instead of directly inferring the HR features, the subnetwork is built to estimate the residual features between LR  and HR features. We denote ${{\mathbf{D}}_{q-1}}$ and ${{\mathbf{D}}_{q}}$ as the input and the output of the $q\text{-th}$ subnetwork, $M$ as the number of MSC modules for stacking, and ${{f}_{q}}$ as the convolution operation of the first convolutional layer. Thus, the output of the $q\text{-th}$ subnetwork is
\begin{equation}
\vspace{-0.1cm}
\begin{aligned}
\label{Eq: eq 3}
{\mathbf{D}}_{q}={} & {{\xi}^{q}} \left( {{\mathbf{D}}_{q-1}} \right)={{G}^{last}}\left( {{G}^{\left( M-1 \right)}}\left( \left[ \begin{matrix}
   {{f}_{q}}\left( {{\mathbf{D}}_{q-1}} \right)  \\
   {{f}_{q}}\left( {{\mathbf{D}}_{q-1}} \right)  \\
\end{matrix} \right] \right) \right)+{{\mathbf{D}}_{q-1}} \\[4pt]
= {} & {{G}^{last}}\left( G\left( \cdots G\left( \left[ \begin{matrix}
   {{f}_{q}}\left( {{\mathbf{D}}_{q-1}} \right)  \\
   {{f}_{q}}\left( {{\mathbf{D}}_{q-1}} \right)  \\
\end{matrix} \right] \right) \cdots \right) \right)+{{\mathbf{D}}_{q-1}},
\end{aligned}
\end{equation}
where $G$ is function for MSC module in last subsection (depicted in Eq.~(\ref{Eq: eq 1})) and ${G}^{last}$ denotes the function of the last MSC module in Eq.~(\ref{Eq: eq 2}).  Since $M\text{-fold}$ operations of $G$ and ${{G}^{last}}$ are performed as well as the residual-features learning is applied, the subnetwork is able to capture different characteristics and representations for inferring the HR features.

\begin{figure}
\captionsetup{belowskip=-18pt}
\centering
\includegraphics[width=0.9\linewidth]{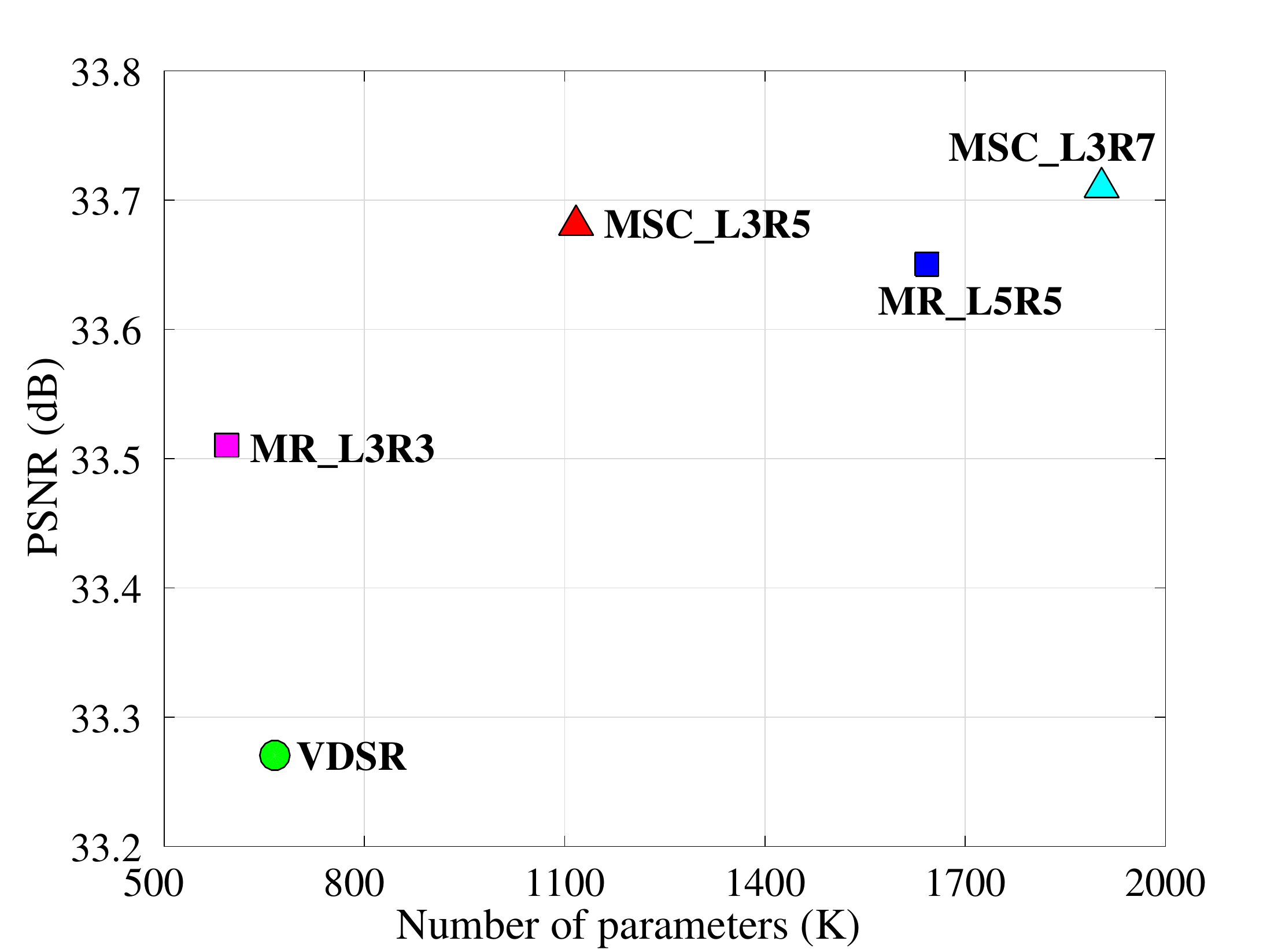}
\caption{The comparisons of PSNRs obtained by the networks in Fig.\ref{fig:Four network sructures} for a scale factor of $3\times$ on the  Set5 dataset. The circle denotes the model of VDSR, while the square and the triangle denote the models with MR blocks and with MSC modules, respectively.}
\label{fig:The comparisons of PSNRs obtained by the networks}
\end{figure}

\subsection{Overall Model with Cascaded Structure}

In order to reconstruct HR features and reduce the difficulty of direct super-resolving the details, we build a cascade of subnetworks to estimate HR features from LR features extracted by the feature extraction network. All subnetworks share the same structure and setting as mentioned in Section III \emph{B} and are jointly trained with simultaneous supervision. It is expected that each stage of subnetwork brings the predictions closer to the ground truth HR features and the cascade progressively minimizes the prediction errors. Then, all the estimated HR features from subnetworks are fed into corresponding reconstruction layers to reconstruct  HR image. The full model is illustrated in Fig.\ref{fig:The architectures} and termed as cascaded multi-scale cross network (CMSC).

Let  $\mathbf{x}$ and  $\hat{\mathbf{y}}$  denote   the  input and output of the CMSC network. And, we utilize a convolutional layer followed by BN and LeakyReLU as a feature extraction network to extract the features from LR input image. The feature extraction network is formulated as below.
\begin{equation}
\label{Eq: eq 4}
{{\mathbf{D}}_{0}}=F\left( \mathbf{x} \right),
\end{equation}
where $F$ denotes the operation of the feature extraction network and ${{\mathbf{D}}_{0}}$ is the extracted features which are then fed into the first stage of subnetwork. Supposing $S$ subnetworks  with  RFL  are cascaded to progressively infer HR features, following the notations in Section III \emph{B}, the process of inference is represented as follows.
\begin{equation}
\label{Eq: eq 5}
{{\mathbf{D}}_{S}}={{\xi }^{S}}\left( {{\xi }^{S-1}}\left( \cdots \left( {{\xi }^{1}}\left( {{\mathbf{D}}_{0}} \right) \right) \cdots \right) \right),
\end{equation}
where ${\xi}^{q}  \left( q=1,2,\cdots ,S \right)$ represents the function for the  $q\text{-th}$  subnetwork, as depicted in Eq.~(\ref{Eq: eq 3}). In order to make the output from each cascaded subnetwork closer to the ground truth HR features, we supervise all predictions from cascaded subnetworks via cascaded-supervision. The output of each subnetwork is  fed into the reconstruction network, where each convolutional layer takes the output of the corresponding stage as its input to reconstruct the corresponding HR residual image, and then $S$ intermediate predictions from the cascaded stages are generated, as illustrated in Fig.\ref{fig:The architectures}(a). Similar to \cite{12KimJ2016CVPR_VDSR, 13KimJ2016CVPR_DRCN, 14TaiY2017CVPR_DRRN,  22TaiY2017ICCV_MemNet}, we also adopt global residual learning (GRL) in our network via adding an identity branch from  the  input to the reconstruction network. Thus, the $q\text{-th} \left( q=1,2,\cdots ,S \right)$ intermediate prediction is
\begin{equation}
\label{Eq: eq 6}
{{\hat{\mathbf{y}}}_{q}}={{R}_{q}}\left( {{\mathbf{D}}_{q}} \right)+\mathbf{x},
\end{equation}
where ${{\mathbf{D}}_{q}}$ is the output features of the $q\text{-th}$ stage and ${{R}_{q}}$ denotes the function for the corresponding reconstruction layer. And then, all of $S$ intermediate predictions are assembled to generate the final output via weighted average.
\begin{equation}
\label{Eq: eq 7}
\hat{\mathbf{y}}=\sum\limits_{q=1}^{S}{{{w}_{q}}}\cdot {{\hat{\mathbf{y}}}_{q}},
\end{equation}
where ${{w}_{q}}$ denotes the weight for the prediction from the $q\text{-th}$ subnetwork. All of the weights in the above equation are learned during training and the final output $\hat{\mathbf{y}}$ is also supervised.

Given a training dataset $\left\{ {{\mathbf{x}}^{(k)}},{{\mathbf{y}}^{(k)}} \right\}_{k=1}^{K}$, where $K$ is the number of training patches and $\left\{ {{\mathbf{x}}^{(k)}},{{\mathbf{y}}^{(k)}} \right\}$  are the $k\text{-th}$ LR and HR patch pairs, the loss function of our model with cascaded-supervision can be formulated as
\begin{equation}
\begin{aligned}
\label{Eq: eq 8}
L\left( \Theta  \right)=\alpha \cdot \sum\limits_{k=1}^{K}{\frac{1}{2K}}{{\Bigl\| {{\mathbf{y}}^{\left( k \right)}}-\sum\limits_{q=1}^{S}{{{w}_{q}}\cdot \hat{\mathbf{y}}_{q}^{(k)}} \Bigr\|^{2}}} \\[4pt]
+\left( 1-\alpha  \right)\cdot \sum\limits_{q=1}^{S}{\sum\limits_{k=1}^{K}{\frac{1}{2SK}}}\Bigl\| {{\mathbf{y}}^{\left( k \right)}}-\hat{\mathbf{y}}_{q}^{(k)} \Bigr\|^{2},
\end{aligned}
\end{equation}
where $ \Theta $ denotes the parameter set. The $\alpha$ balances the losses on the final output and on the intermediate outputs, and is empirically set as $\frac{2}{S+2}$.

When the depth of a network is defined as the longest path from the input to the  output, the depth of our overall model can be calculated by
\begin{equation}
\label{Eq: eq 9}
Depth=\left( M\cdot 2+1 \right)\cdot S+2,
\end{equation}
where $S$ and $M$ denote the number of cascaded subnetworks and the number of MSC modules in each subnetwork. The $2$ multiplied by $M$ represents two convolutional layers contained in a branch of MSC module, and the $1$ in the parentheses indicates the first convolutional layer in each subnetwork. Besides, the $2$ at the end of equation represents two convolutional layers in feature extraction network and in reconstruction network respectively.

\begin{figure}[t]
\captionsetup[subfigure]{farskip = 0pt}
\captionsetup{belowskip=-16pt}
    \small
    \begin{minipage}[b]{1\linewidth}
        \centering
        \subfloat[The network with plain structure]{
            \centering
            \includegraphics[width=1\linewidth]{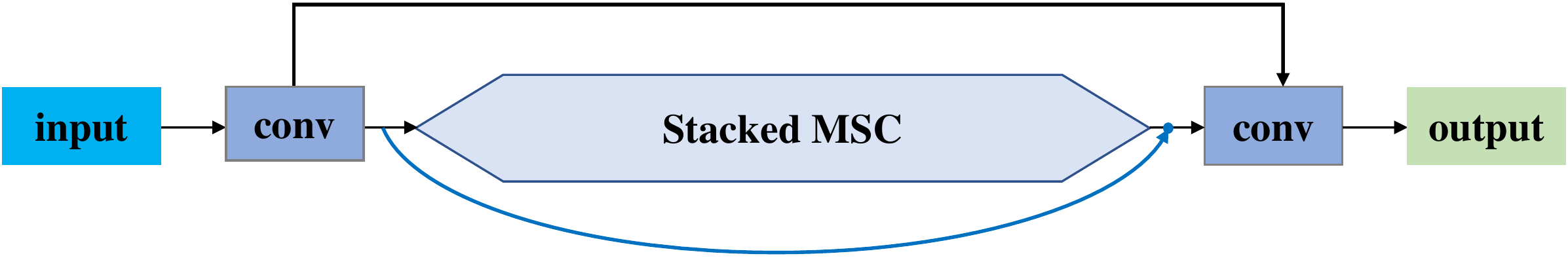}
        }\\
       \vspace{0.2cm}
        \subfloat[The network with cascaded structure]{
            \centering
            \includegraphics[width=1\linewidth]{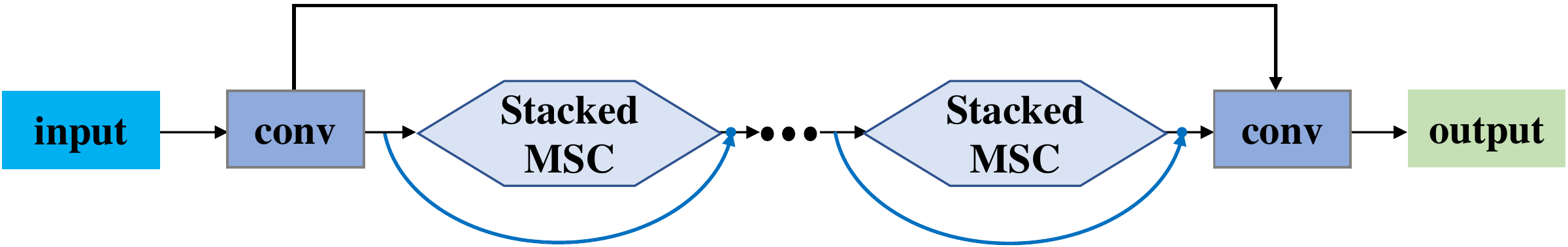}
        }
        \end{minipage}
\caption{The plain structure and the cascaded structure for the networks containing 15 stacked MSC modules. (a) The network with plain structure, which consists of two convolutional layers for features representation and image reconstruction respectively and 15 stacked MSC modules in chain mode for inferring HR features. (b) The network with cascaded structure, where three subnetworks with five stacked MSC modules are cascaded for inferring the HR features in a coarse-to-fine manner.}
\label{fig:The plain structure and the cascaded structure}
\end{figure}

\section{Experiments and Analysis}

\subsection{Benchmarks}
We conduct comparison studies on widely used datasets, Set5 \cite{40Bevilacqua2012BMVC}, Set14 \cite{41Zeyde2010ICCS}, BSD100 \cite{42ArbelaezP2011TPAMI}  and Urban100 \cite{43Huang2015CVPRSelfExp}, which contain 5, 14, 100 and 100 images respectively. We use a training set of 291 images for benchmark with other methods, which consists of 91 images from Yang \etal \cite{5Yang2010TIP}  and 200 images from the training set of BSD300 dataset  \cite{42ArbelaezP2011TPAMI}. For results in the section of model analysis, 91 images \cite{5Yang2010TIP}  are used to train network fast. In addition, data augmentation is used, which includes flipping, downscaling with the scales of $0.7$ and $0.5$, and rotating by ${{90}^{\circ }}$, ${{180}^{\circ }}$ and ${{270}^{\circ }}$.

We use the peak signal-to-noise ratio (PSNR), the structural similarity (SSIM) \cite{47Wang2004TIPSSIM} index and information fidelity criterion (IFC) \cite{44Sheikh2005TIPIFC} as metrics for evaluation. Since the reconstruction is performed on the Y-channel in YCbCr color space, all the criteria are calculated on the Y-channel of images after pixels near image boundary are removed.

\subsection{Implementation Details}
Given the HR image, the input LR image is generated by first downsampling and then upsampling with bicubic interpolation to a certain scale and has the same size as the HR image. By following \cite{12KimJ2016CVPR_VDSR}, we only train a single model for all different scales, including $2\times$, $3\times$ and $4\times$. The LR and HR pairs of all different scales are combined into one training dataset.

We split training images into $41\times41$  sub-images with no overlap and set the mini-batch size to 32 for stochastic gradient descent (SGD) solver. And, we set momentum parameter to $0.9$ and the weight decay to ${{10}^{-4}}$. The initial learning rate is initialized to $0.1$ and then decreased by a factor of 10  for every 10 epochs. To suppress the vanishing and the exploding gradients, we clip individual gradients to a certain range $\left[ -\eta , \eta  \right]$, where $\eta $ is set to $0.4$. We find that our convergent  procedure is fast and our model gets convergence after about 50 epochs.

In our CMSC network, each convolutional layer has 64 filters and is followed by BN and LeakyReLU with a negative slope of $0.2$. The kernel size of the convolutional layers is set to $3\times3$ except the convolutional layers in MSC modules, of which kernel size is determined according to the experimental analysis in Section IV \emph{C}. For weight initialization in all convolutional layers, we applied the same way as in He \etal  \cite{45He2015CVPRInitial}.  And, we apply PyTorch on a NVIDIA Titan X Pascal GPU (12G Memory) for model training and testing.

\begin{table}[b]
\vspace{-0.2cm}
\captionsetup{justification=centering}
\caption{\textsc{\\ Study on the Effect of Cascaded Structure. \\ Average PSNRs  for a  Scale  Factor of  $2\times$ Are Reported. \\ {\bf \textsc{Fontbold}} Indicates the
Best Performance.}}
\footnotesize
\centering
\begin{tabular}{{p{2cm}<{\centering} p{1cm}<{\centering} p{1cm}<{\centering}
p{1cm}<{\centering} p{1.2cm}<{\centering}}}
\hline
\hline
    Dataset & \textsc{Set5} & \textsc{Set14} & \textsc{BSD100} & \textsc{Urban100}\\
\hline
    PMSC & 37.569 & 33.094 & 31.891 & 30.846\\
\hline
    CMSC\_NS & {\bf 37.609} & {\bf 33.121} & {\bf 31.894} & {\bf 30.889}\\
\hline
\hline
\end{tabular}
\label{tab:Study on the effect of cascaded structure}
\end{table}

\subsection{Model Analysis}
In this section, the designs and the contributions of different components of our model are analyzed via the experiments, including MSC module, cascaded structure, residual-features learning, cascaded-supervision  and different reconstruction layers utilization. For all experiments in this section, we use the models trained on 91 images \cite{5Yang2010TIP}  for faster training.

\subsubsection{Multi-scale cross module}
To design multi-scale cross (MSC) module for fusing multiple levels information as well as to demonstrate the superiority of designed MSC module, we build four modules for comparison. Two merge-and-run blocks (MR) with different filter sizes are shown in Fig.\ref{fig:Four network sructures}(a): (I) MR\_L3R3, where four convolutional layers in two assembled residual branches have 64 filters of the size $3\times3$, (II) MR\_L5R5, in which each convolutional layer has 64 filters with the size of $5\times5$. In addition, Fig.\ref{fig:Four network sructures}(a) depicts two MSC modules which are utilized to fuse information from different receptive fields: (III) MSC\_L3R5, in which one residual branch contains two stacked convolutional layers with 64 filters of the size $3\times3$, and another residual branch uses the same number of filters but with the size of $5\times5$ for the convolutional layers, (IV) MSC\_L3R7, which has the similar structure with the MSC\_L3R5, but stacks two convolutional layers with the filter kernel size of $7\times7$ in one residual branch (orange rectangles in Fig.\ref{fig:Four network sructures}(a)). We stack these different modules to construct the corresponding plain networks with global residual learning (GRL) (Fig.\ref{fig:Four network sructures}(b)) for comparing SR performances. As illustrated in Fig.\ref{fig:Four network sructures}(b), the built basic network is composed of a convolutional layer, four stacked modules and another convolutional layer for reconstructing the residual image. By applying four modules (Fig.\ref{fig:Four network sructures}(a)) into the basic structure in Fig.\ref{fig:Four network sructures}(b) respectively, we obtain four networks which are named according to their containing modules (i.e., MR\_L3R3, MR\_L5R5, MSC\_L3R5 and MSC\_L3R7). In addition, we compare these four networks with VDSR \cite{12KimJ2016CVPR_VDSR}  (illustrated in Fig.\ref{fig:Four network sructures}(c)) which has 20 stacked convolutional layers. We apply the trained models of these networks on  the  Set5 dataset  and then illustrate the PSNRs of these structures for SR with a scale factor of $3\times$ in Fig.\ref{fig:The comparisons of PSNRs obtained by the networks}. One can see that, by introducing interactions between branches, MR\_L3R3 network with fewer parameters and fewer layers achieves higher PSNR than VDSR, which manifests the effectiveness of the merge-and-run mapping for SR. On the other hand, our proposed MSC\_L3R5 performs better than MR\_L5R5 but contains fewer parameters, which suggests that the gains of our MSC module over MR block (MR\_L3R3 and MR\_L5R5) are not only from the larger receptive field but also from multi-scale complementary information fusing and richer representation. Among these models, MSC\_L3R7 achieves the best performance. Considering both the performance and the number of parameters, we adopt the MSC\_L3R5 as our MSC module for the next experiments.

\subsubsection{Cascaded structure}
To validate the effectiveness of the cascaded structure, we use fifteen MSC modules to build a plain structure and a cascaded structure for comparing, which are shown in Fig.\ref{fig:The plain structure and the cascaded structure}. For the plain structure, denoted as PMSC, we stack fifteen MSC modules to reconstruct the HR features and also apply the residual-features learning (RFL) as well as  the global residual learning (GRL) in this structure. For cascaded structure, we utilize five MSC modules in each subnetwork and adopt three subnetworks for three stages of cascade. For fair comparison, cascaded-supervision and predictions ensemble are excluded from the cascaded structure, which is denoted as CMSC\_NS. Both the plain structure and the cascaded structure use one convolutional layer for the features extraction layer and for the reconstruction layer respectively. TABLE~\ref{tab:Study on the effect of cascaded structure}  presents the SR results on four benchmark datasets, including Set5, Set14, BSD100 and Urban100. From the results, it is seen that the cascaded structure  (CMSC\_NS)  achieves moderate performance improvements over the plain structure. Therefore, the image SR can benefit from the cascaded structure.

\begin{table}[t]
\captionsetup{belowskip=-0pt,justification=centering}
\caption{\textsc{\\ Study on the Effects of Residual-Features Learning, Cascaded-Supervision and  Different Reconstruction Layers Utilization.  Average PSNRs/SSIMs on the Set5 Dataset Are Reported.  {\bf \textsc{Fontbold}} Indicates the Best Performance.}}
\footnotesize
\centering
\begin{tabular}{c c c c c}
\hline
\hline
\multirow{2}{*}{Scale} & CMSC\_NRS & CMSC\_NS & CMSC\_SR & CMSC \\
	& PSNR/SSIM & PSNR/SSIM & PSNR/SSIM & PSNR/SSIM\\
\hline
    $2\times$ & 37.59/0.9592 & 37.61/0.9593 & 37.62/0.9594 & {\bf37.64}/{\bf0.9595}\\
\hline
    $3\times$ & 33.94/0.9238 & 33.95/0.9238 & 33.98/0.9242 & {\bf34.03}/{\bf0.9246}\\
\hline
    $4\times$ & 31.55/0.8870 & 31.58/0.8874 & 31.61/0.8879 & {\bf31.65}/{\bf0.8887}\\
\hline
\hline
\end{tabular}
\label{tab:Study on the effect of residual-features learning}
\end{table}

\begin{figure}
\captionsetup[subfigure]{farskip = 0pt}
\captionsetup{belowskip=-16pt}
    \small
    \begin{minipage}[b]{1\linewidth}
        \centering
        \subfloat[]{
            \centering
            \includegraphics[width=0.5\linewidth]{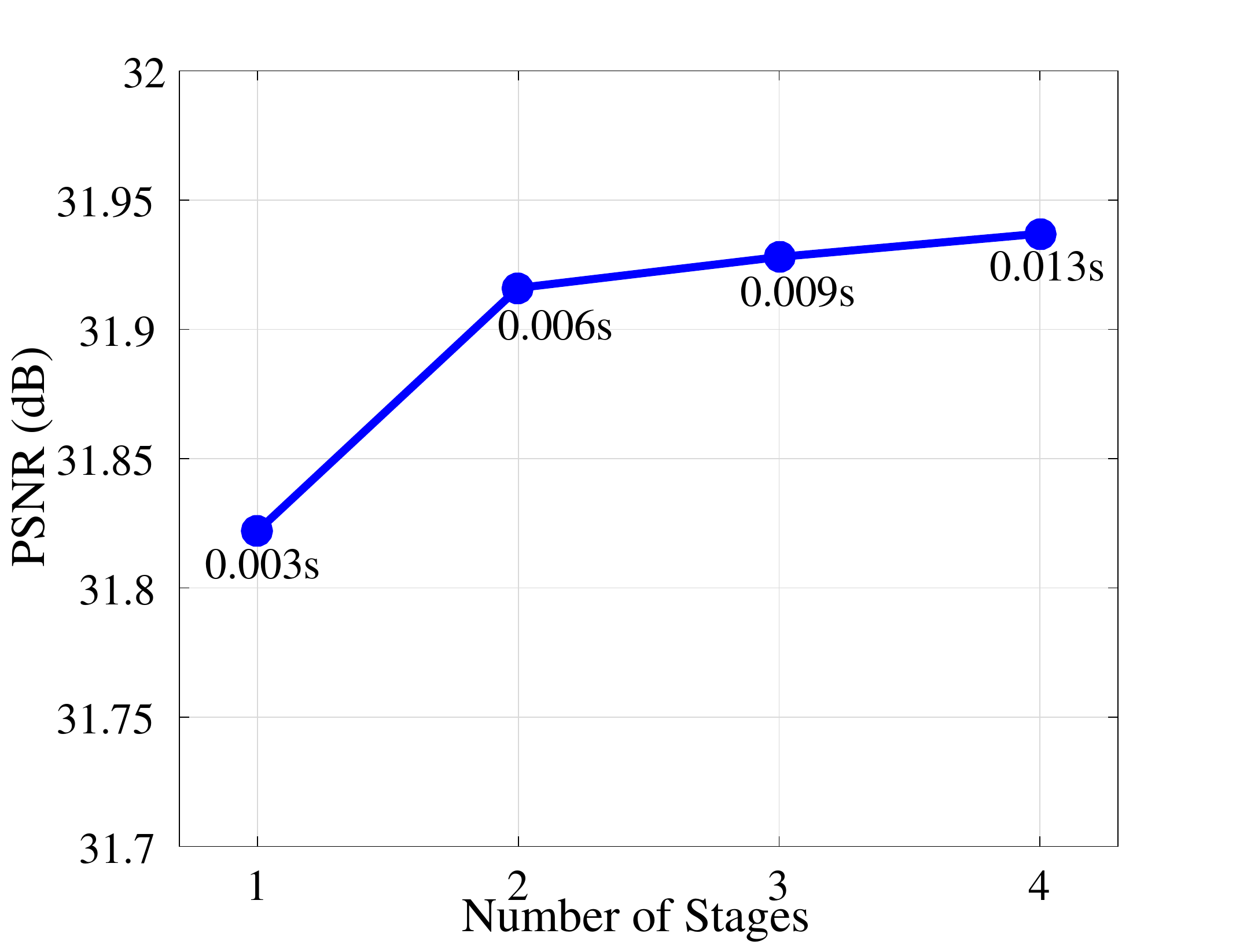}
        }
        \subfloat[]{
            \centering
            \includegraphics[width=0.5\linewidth]{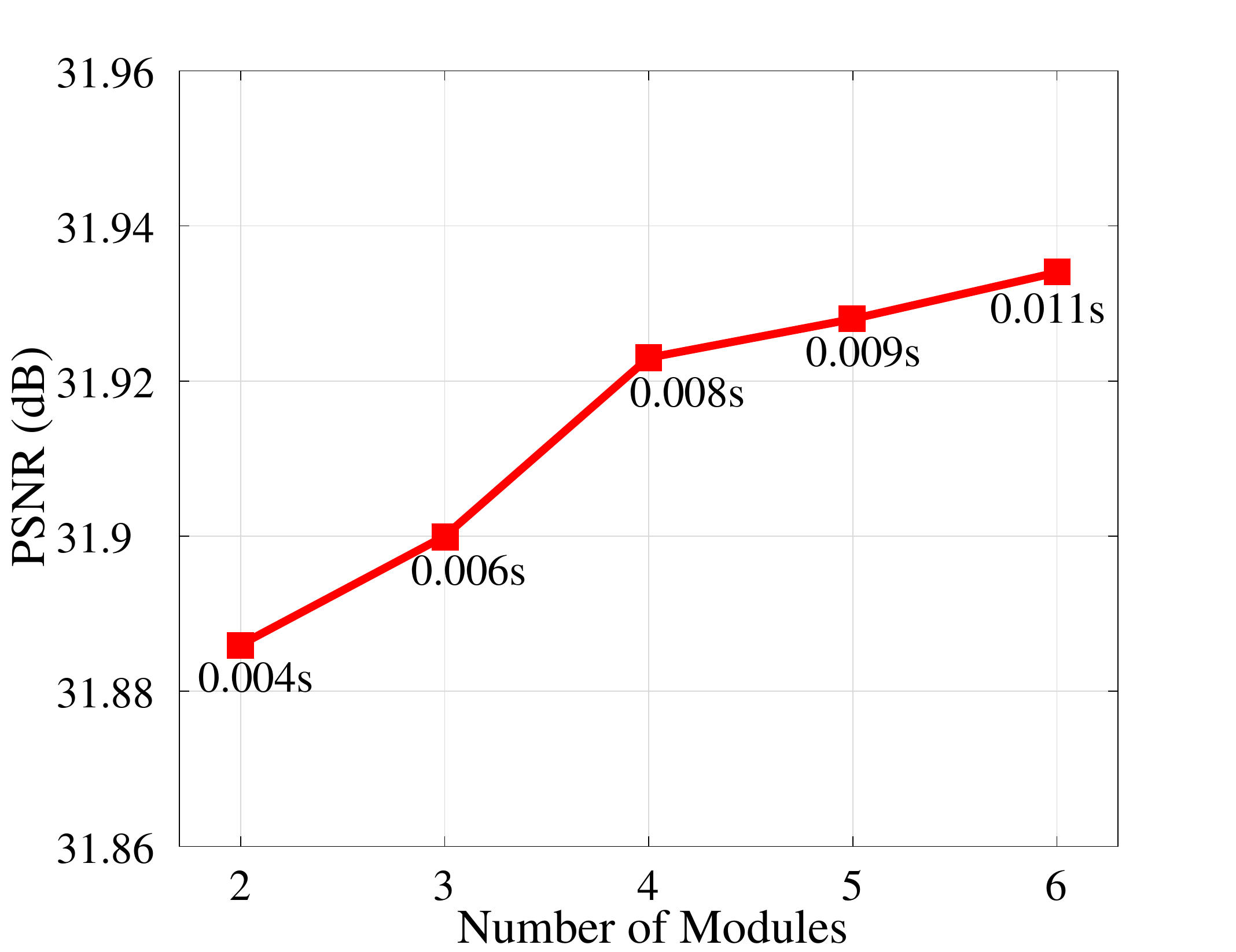}
        }
        \end{minipage}
\caption{ (a) PSNR performance versus the number of cascaded stages ($S$). (b) PSNR performance versus the number of modules ($M$) in each stage. The numbers marked on the side of the curves represent the execution time in seconds. The tests are conducted for a scale factor of $2\times$ on the dataset of BSD100.}
\label{fig:PSNR performance versus the number of cascaded stages}
\end{figure}

\begin{table*}[t]
\captionsetup{belowskip=-8pt,justification=centering}
\caption{\textsc{ \\ Quantitative Evaluations of State-of-the-art SR Methods.\\
The Average PSNRs/SSIMs/IFCs for Scale Factors of $2\times$, $3\times$ and $4\times$ Are Reported. \\
{\bf\textsc{Fontbold}} Indicates the Best Performance and {\underline{\textsc{Underline}}} Indicates the Second-best Performance.
}}
\footnotesize
\centering
\begin{tabular}{{c|c|c c c|c c c|c c c|c c c}}
\hline
\hline
\multirow{2}{*}{Scale} & \multirow{2}{*}{Method} & \multicolumn{3}{c|}{\textsc{Set5}} & \multicolumn{3}{c|}{\textsc{Set14}} & \multicolumn{3}{c|}{\textsc{BSD100}} & \multicolumn{3}{c}{\textsc{Urban100}} \\
	& & PSNR & SSIM & IFC & PSNR & SSIM & IFC & PSNR & SSIM & IFC & PSNR & SSIM & IFC\\
\hline
\multirow{12}*{$2\times$}
	& Bicubic
    & 33.68 & 0.9304 & 6.2821 & 30.24 & 0.8691 & 6.3035 & 29.56 & 0.8435 & 5.9201 & 26.88 & 0.8405 & 6.6493\\
	& A+~\cite{7Timofte2014ACCV}
    & 36.54 & 0.9544 & 8.4772 & 32.28 & 0.9056 & 8.1395 & 31.21 & 0.8863 & 7.3556 & 29.20 & 0.8938 & 8.4201\\
    & SelfExSR~\cite{43Huang2015CVPRSelfExp}
    & 36.50 & 0.9536 & 7.8240 & 32.22 & 0.9034 & 7.5970 & 31.17 & 0.8853 & 6.8356 & 29.52 & 0.8965 & 7.9437\\
    & SRCNN~\cite{11Dong2016TPAMI}
    & 36.66 & 0.9542 & 8.0356 & 32.45 & 0.9067 & 7.7844 & 31.36 & 0.8879 & 7.1260 & 29.51 & 0.8946 & 7.9958\\
    & FSRCNN~\cite{46Dong2016ECCVFSRCNN}
    & 36.98 & 0.9556 & 7.8118 & 32.62 & 0.9087 & 7.5656 & 31.50 & 0.8904 & 6.8622 & 29.85 & 0.9009 & 8.0255\\
    & VDSR~\cite{12KimJ2016CVPR_VDSR}
    & 37.53 & 0.9587 & 8.5801 & 33.05 & 0.9127 & 8.1594 & 31.90 & 0.8960 & 7.4942 & 30.77 & 0.9141 & 8.6288\\
    & DRCN~\cite{13KimJ2016CVPR_DRCN}
    & 37.63 & 0.9588 & 8.7828 & 33.06 & 0.9121 & 8.3700 & 31.85 & 0.8942 & 7.5766 & 30.76 & 0.9133 & 8.9590\\
    & LapSRN~\cite{29LaiWS2017CVPR_LapSRN}
    & 37.52 & 0.9591 & 9.0133 & 32.99 & 0.9124 & 8.5085 & 31.80 & 0.8949 & 7.7192 & 30.41 & 0.9101 & 8.9227\\
    & DRRN~\cite{14TaiY2017CVPR_DRRN}
    & 37.74 & 0.9591 & 8.6698 & 33.23 & 0.9136 & 8.3178 & 32.05 & 0.8973 & 7.5128 & 31.23 & 0.9188 & 8.9200\\
    & MemNet~\cite{22TaiY2017ICCV_MemNet}
    & {\underline{37.78}} & 0.9597 & 8.8503 & 33.28 & 0.9142 & 8.4682 & {\underline{32.08}} & 0.8978 & 7.6650  & {\underline{31.31}} & {\underline{0.9195}} & 9.1221\\
    & CMSC\_SR (ours)
    & 37.76 & {\underline{0.9600}} & {\underline{9.1463}} & {\underline{33.29}} & {\underline{0.9148}} & {\underline{8.7113}} & 32.07 & {\underline{0.8984}} & {\underline{7.9785}} & 31.21 & 0.9191 & {\underline{9.4034}}\\
    & CMSC (ours)
    & {\bf 37.89} & {\bf 0.9605} & {\bf 9.3922} & {\bf 33.41} & {\bf 0.9153} & {\bf 8.8932} & {\bf 32.15} & {\bf 0.8992} & {\bf 8.1380} & {\bf 31.47} & {\bf 0.9220} & {\bf 9.6666}\\
\hline
\multirow{12}*{$3\times$}
	& Bicubic
    & 30.40 & 0.8686 & 3.6160 & 27.54 & 0.7741 & 3.5346 & 27.21 & 0.7389 & 3.2339 & 24.46 & 0.7349 & 3.7759\\
	& A+~\cite{7Timofte2014ACCV}
    & 32.58 & 0.9088 & 4.9285 & 29.13 & 0.8188 & 4.5348 & 28.29 & 0.7835 & 3.9968 & 26.03 & 0.7973 & 4.8631\\
    & SelfExSR~\cite{43Huang2015CVPRSelfExp}
    & 32.64 & 0.9097 & 4.7550 & 29.15 & 0.8196 & 4.3704 & 28.29 & 0.7840 & 3.8003 & 26.46 & 0.8090 & 4.8397\\
    & SRCNN~\cite{11Dong2016TPAMI}
    & 32.75 & 0.9090 & 4.6582 & 29.29 & 0.8215 & 4.3377 & 28.41 & 0.7863 & 3.8435 & 26.24 & 0.7991 & 4.5790\\
    & FSRCNN~\cite{46Dong2016ECCVFSRCNN}
    & 33.16 & 0.9140 & 4.9640 & 29.42 & 0.8242 & 4.5494 & 28.52 & 0.7893 & 4.0302 & 26.41 & 0.8064 & 4.8419\\
    & VDSR~\cite{12KimJ2016CVPR_VDSR}
    & 33.66 & 0.9213 & 5.2029 & 29.78 & 0.8318 & 4.6906 & 28.83 & 0.7976 & 4.1514 & 27.14 & 0.8279 & 5.1594\\
    & DRCN~\cite{13KimJ2016CVPR_DRCN}
    & 33.82 & 0.9226 & 5.3363 & 29.77 & 0.8314 & 4.7816 & 28.80 & 0.7963 & 4.1839 & 27.15 & 0.8277 & 5.3145\\
    & LapSRN~\cite{29LaiWS2017CVPR_LapSRN}
    & 33.82 & 0.9227 & 5.1946 & 29.79 & 0.8320 & 4.6635 & 28.82 & 0.7973 & 4.0581 & 27.07 & 0.8271 & 5.1649\\
    & DRRN~\cite{14TaiY2017CVPR_DRRN}
    & 34.03 & 0.9244 & 5.3962 & 29.96 & 0.8349 & 4.8773 & 28.95 & 0.8004 & 4.2350 & 27.53 & {\underline{0.8377}} & 5.4496\\
    & MemNet~\cite{22TaiY2017ICCV_MemNet}
    & 34.09 & 0.9248 & {\underline{5.5027}} & {\underline{30.00}} & 0.8350 & 4.9584 & {\underline{28.96}} & 0.8001 & 4.3002 & {\underline{27.56}} & 0.8376 & 5.5727\\
    & CMSC\_SR (ours)
    & {\underline{34.10}} & {\underline{0.9254}} & 5.4914 & {\underline{30.00}} & {\underline{0.8355}} & {\underline{4.9726}} & {\underline{28.96}} & {\underline{0.8012}} & {\underline{4.3552}} & 27.52 & 0.8369 & {\underline{5.5974}}\\
    & CMSC (ours)
    & {\bf 34.24} & {\bf 0.9266} & {\bf 5.6612} & {\bf 30.09} & {\bf 0.8371} & {\bf 5.1019} & {\bf 29.01} & {\bf 0.8024} & {\bf 4.4427} & {\bf 27.69} & {\bf 0.8411} & {\bf 5.7819}\\
\hline
\multirow{12}*{$4\times$}
	& Bicubic
    & 28.43 & 0.8109 & 2.3425 & 26.00 & 0.7023 & 2.2593 & 25.96 & 0.6678 & 2.0210 & 23.14 & 0.6574 & 2.4459\\
	& A+~\cite{7Timofte2014ACCV}
    & 30.28 & 0.8603 & 3.2483 & 27.32 & 0.7491 & 2.9616 & 26.82 & 0.7087 & 2.5509 & 24.32 & 0.7183 & 3.2080\\
    & SelfExSR~\cite{43Huang2015CVPRSelfExp}
    & 30.30 & 0.8620 & 3.1812 & 27.38 & 0.7516 & 2.8908 & 26.84 & 0.7106 & 2.4601 & 24.80 & 0.7377 & 3.3148\\
    & SRCNN~\cite{11Dong2016TPAMI}
    & 30.48 & 0.8628 & 2.9910 & 27.50 & 0.7513 & 2.7506 & 26.90 & 0.7103 & 2.3961 & 24.52 & 0.7226 & 2.9632\\
    & FSRCNN~\cite{46Dong2016ECCVFSRCNN}
    & 30.70 & 0.8657 & 2.9862 & 27.59 & 0.7535 & 2.7068 & 26.96 & 0.7128 & 2.3297 & 24.60 & 0.7258 & 2.8945\\
    & VDSR~\cite{12KimJ2016CVPR_VDSR}
    & 31.35 & 0.8838 & 3.5419 & 28.02 & 0.7678 & 3.1065 & 27.29 & 0.7252 & 2.6785 & 25.18 & 0.7525 & 3.4623\\
    & DRCN~\cite{13KimJ2016CVPR_DRCN}
    & 31.53 & 0.8854 & 3.5426 & 28.03 & 0.7673 & 3.0978 & 27.24 & 0.7233 & 2.6328 & 25.14 & 0.7511 & 3.4654\\
    & LapSRN~\cite{29LaiWS2017CVPR_LapSRN}
    & 31.54 & 0.8866 & 3.5596 & 28.09 & 0.7694 & 3.1462 & 27.32 & 0.7264 & 2.6771 & 25.21 & 0.7553 & 3.5305\\
    & DRRN~\cite{14TaiY2017CVPR_DRRN}
    & 31.68 & 0.8888 & 3.7022 & 28.21 & 0.7720 & 3.2518 & 27.38 & 0.7284 & 2.7459 & 25.44 & {\underline{0.7638}} & 3.6741\\
    & MemNet~\cite{22TaiY2017ICCV_MemNet}
    & 31.74 & 0.8893 & {\underline{3.7860}} & 28.26 & 0.7723 & 3.3079 & 27.40 & 0.7281 & 2.7784 & {\underline{25.50}} & 0.7630 & 3.7860\\
    & CMSC\_SR (ours)
    & {\underline{31.77}} & {\underline{0.8903}} & 3.7789 & {\underline{28.27}} & {\underline{0.7733}} & {\underline{3.3261}} & {\underline{27.41}} & {\underline{0.7296}} & {\underline{2.8243}} & 25.49 & 0.7637 & {\underline{3.8086}}\\
    & CMSC (ours)
    & {\bf 31.91} & {\bf 0.8923} & {\bf 3.8758} & {\bf 28.35} & {\bf 0.7751} & {\bf 3.4056} & {\bf 27.46} & {\bf 0.7308} & {\bf 2.8767} & {\bf 25.64} & {\bf 0.7692} & {\bf 3.9437}\\
\hline
\hline
\end{tabular}
\label{tab:Quantitative evaluations of state-of-the-art SR methods}
\end{table*}

\begin{figure*}
\vspace{-0.4cm}
\centering
\captionsetup{belowskip=-8pt,aboveskip=-0.005pt,width=0.99\linewidth}
\includegraphics[width=0.99\linewidth]{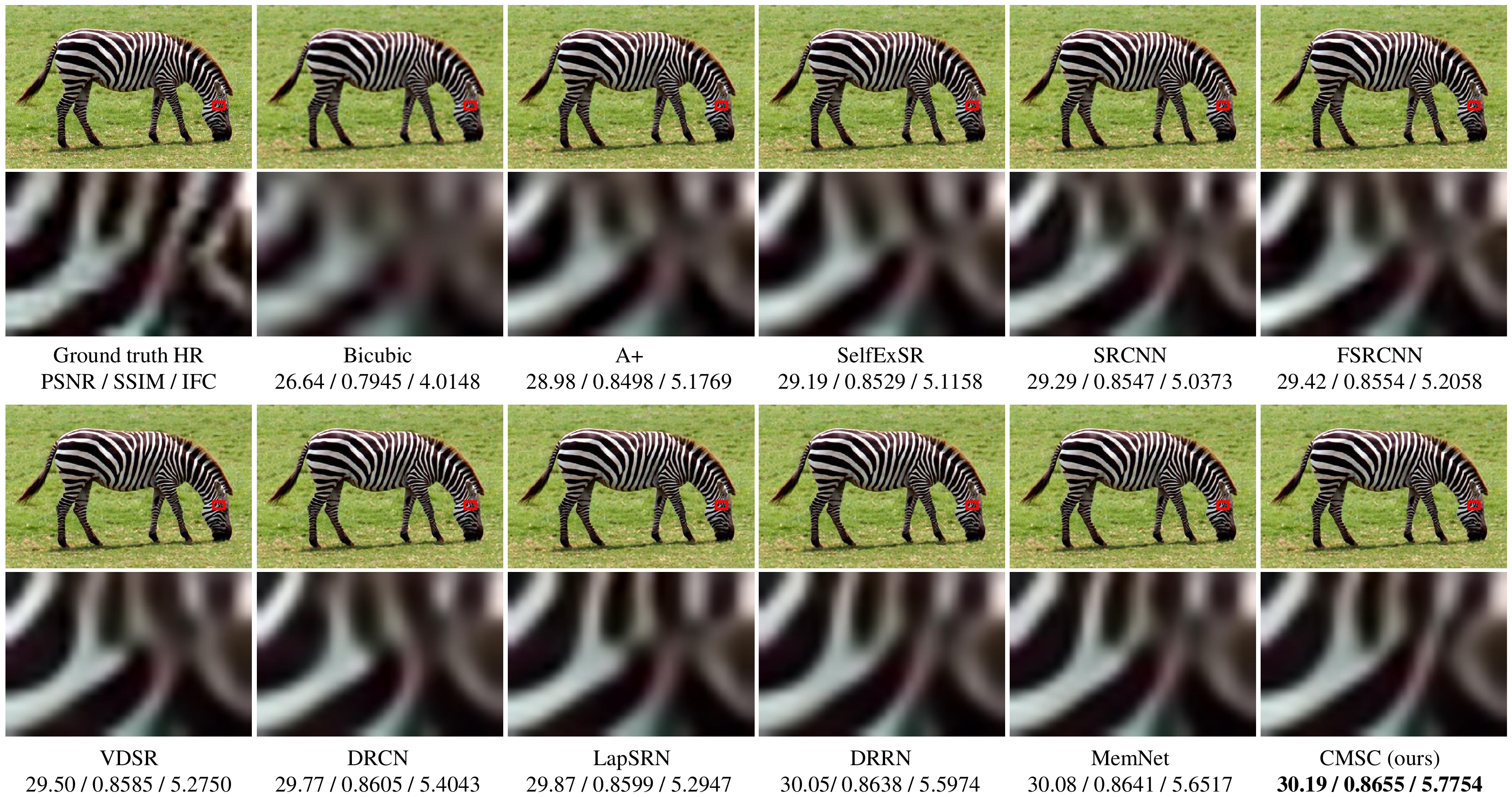}
\caption{Visual evaluation for a scale factor of $3\times$ on  the  ``zebra''  image from  Set14. The CMSC accurately reconstructs zebra stripes while others generate blurry results with severe distortions.}
\label{fig: x3 zebra}
\end{figure*}

\begin{figure*}
\vspace{-0.2cm}
\centering
\captionsetup{belowskip=-14pt,aboveskip=-0.005pt,width=0.99\linewidth}
\includegraphics[width=0.99\linewidth]{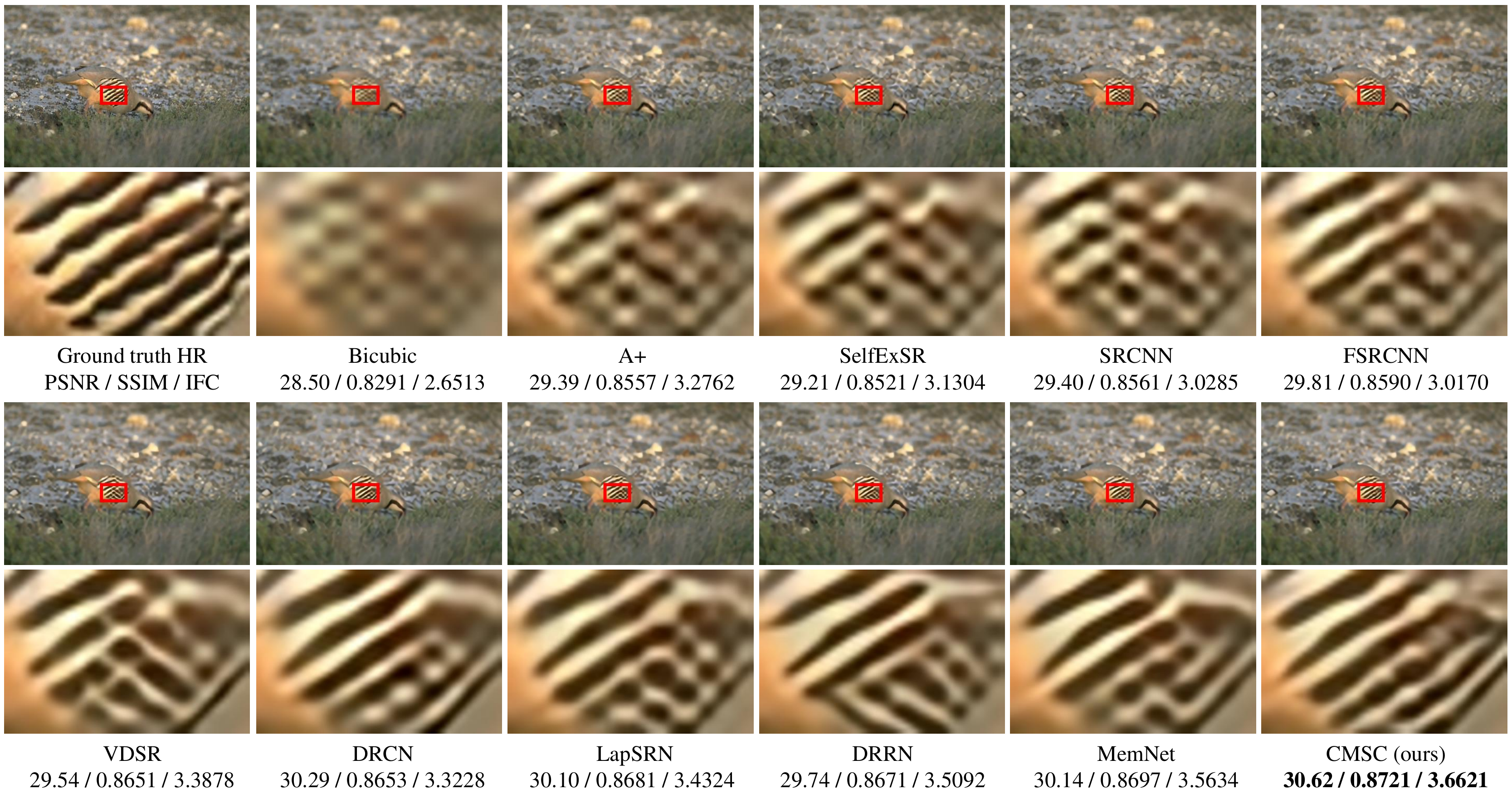}
\caption{Visual evaluation for a scale factor of $4\times$ on  the ``8023'' image  from  BSD100. Only the CMSC correctly reconstructs the textures on the wing.}
\label{fig: x4 8023}
\end{figure*}

\subsubsection{Residual-features learning, cascaded-supervision and different reconstruction layers utilization}
To study contributions of residual-features learning (RFL),  cascaded-supervision and different reconstruction layers utilization  for boosting SR performance, we rebuild three models for comparison besides our final model (CMSC), which are termed as CMSC\_NRS, CMSC\_NS and CMSC\_SR respectively. For the CMSC\_NRS,   the identity branch between the beginning and the end of each cascaded subnetwork (blue  curves in Fig.\ref{fig:The architectures}(b)) is removed from the CMSC, and  the  multiple supervisions are also excluded. Thus,  by directly feeding the output from the last cascaded stage  into the reconstruction network,  the final prediction is obtained. Based on the CMSC\_NRS, we recover the RFL in each subnetwork to obtain the CMSC\_NS in which the cascaded-supervision is still not applied. The difference between the CMSC\_SR and the CMSC is that all subnetworks in CMSC\_SR share the same reconstruction layer to obtain intermediate predictions while the subnetworks in CMSC own their respective reconstruction layers. The four networks have the same number of cascaded subnetworks $\left( S=3 \right)$ and the same number of MSC modules $\left( M=5 \right)$ in each subnetwork. TABLE~\ref{tab:Study on the effect of residual-features learning} shows the SR performances of four models in terms of PSNR and SSIM on  the Set5 dataset for three scale factors of $2\times$, $3\times$ and $4\times$. We can see that both residual-features learning and cascaded-supervision make contributions to improving the SR performance. Further, the CMSC achieves better performance than CMSC\_SR with very few parameters increase, which manifests that applying different reconstruction layers for different cascaded stages can further boost SR performance.

\subsubsection{The number of stages ($S$) and the number of modules ($M$)}
The capacity of the CMSC is mainly determined by two parameters: the number of subnetworks ($S$) for cascaded and the number of MSC modules ($M$) in each subnetwork. In this subsection, we test the effects of the two parameters on image SR. Firstly, we fix the parameter of $M$ to $5$ and change the parameter of $S$ from $1$ to $4$. Fig.\ref{fig:PSNR performance versus the number of cascaded stages}(a) shows the curve of the PSNR performance versus the parameter $S$ on the dataset of BSD100 for a scale factor of $2\times$, and the corresponding average execution time in seconds is marked on the side of the curve. We can see that the performance is improved gradually with the increase in the number of stages but at the expense of increased computational cost.

Next, the parameter $S$ is fix to $3$ and the parameter $M$ in each stage is increased from $2$ to $6$. The curve of the PSNR performance versus the parameter $M$ is illustrated in Fig.\ref{fig:PSNR performance versus the number of cascaded stages}(b). When the more MSC modules are contained in subnetwork, the network gets deeper. Therefore, the curve in Fig.\ref{fig:PSNR performance versus the number of cascaded stages}(b) illustrates that the deeper network still achieves the better performance but with more execution time. To strike a balance between performance and speed, we choose $M=5$ and $S=3$ for our CMSC model, the depth of which is 35 according to Eq.~(\ref{Eq: eq 9}).

\begin{figure*}
\vspace{-0.4cm}
\centering
\captionsetup{belowskip=-14pt,aboveskip=-0.005pt,width=0.99\linewidth}
\includegraphics[width=0.99\linewidth]{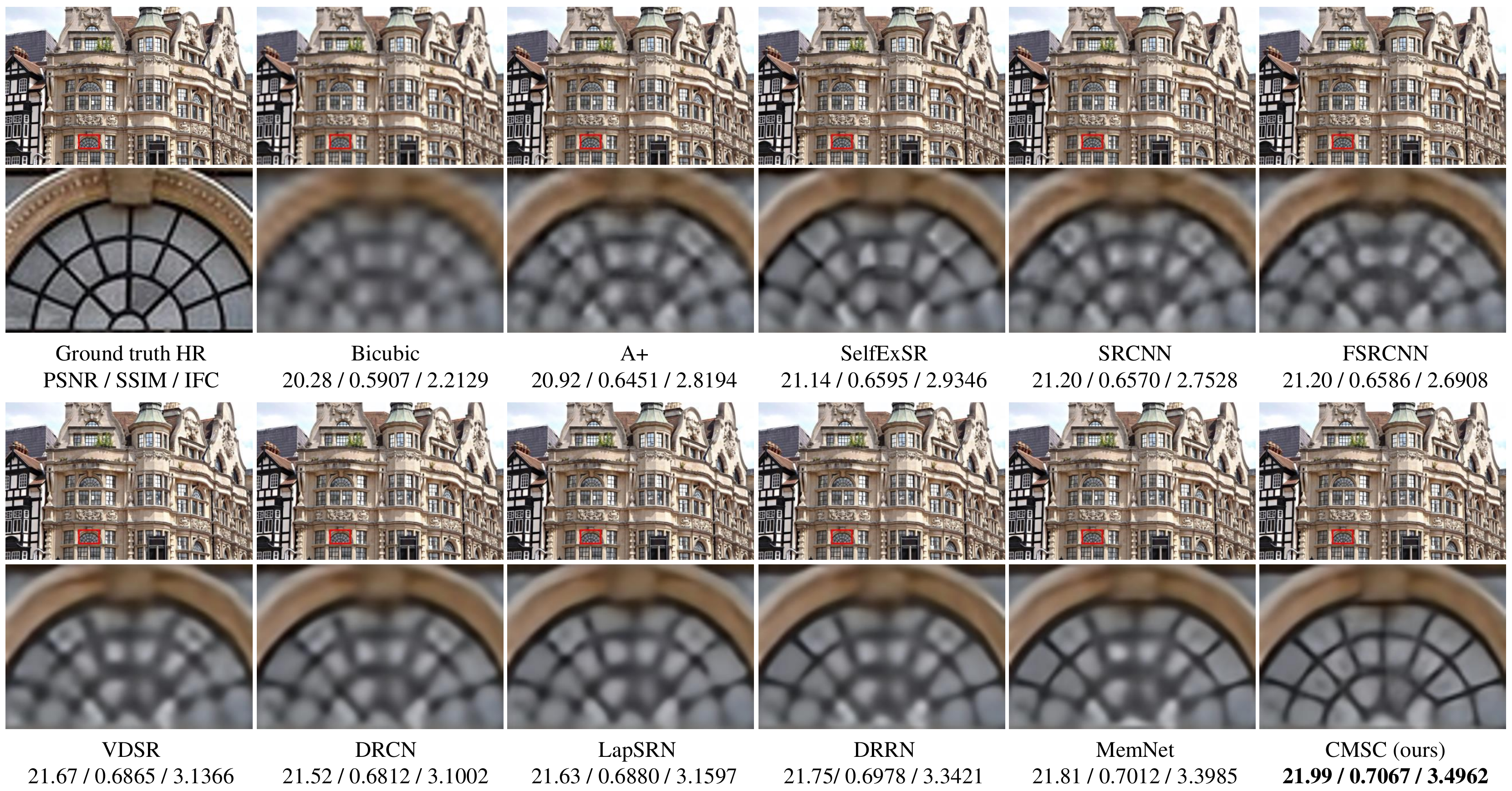}
\caption{Visual evaluation for a scale factor of $4\times$ on the  ``img053''  image  from  Urban100. The contours of window are cleaner in result of CMSC than in other results.}
\label{fig: x4 img053}
\end{figure*}

\subsection{Comparisons With the State-of-the-arts}
\vspace{-0.1cm}
To illustrate the effectiveness of the proposed CMSC model, several state-of-the-art single image SR methods, including A+ \cite{7Timofte2014ACCV}, SelfExSR \cite{43Huang2015CVPRSelfExp}, SRCNN \cite{11Dong2016TPAMI}, FSRCNN \cite{46Dong2016ECCVFSRCNN}, VDSR \cite{12KimJ2016CVPR_VDSR}, DRCN \cite{13KimJ2016CVPR_DRCN}, LapSRN \cite{29LaiWS2017CVPR_LapSRN}, DRRN  \cite{14TaiY2017CVPR_DRRN}  and MemNet  \cite{22TaiY2017ICCV_MemNet}, are compared in terms of quantitative evaluation, visual quality and execution time. For comparison, we also construct the CMSC\_SR model which has the same parameters of $S$ ($S=3$) and $M$ ($M=5$) as the CMSC model but enables the reconstruction network sharing among all stages of subnetworks, similar to DRCN  \cite{13KimJ2016CVPR_DRCN}  and MemNet  \cite{22TaiY2017ICCV_MemNet}. All methods are only applied to the luminance channel of an image while bicubic interpolation is utilized to the color components.

The quantitative evaluations on the four benchmark datasets for three scale factors ($2\times$, $3\times$, $4\times$) are summarized in TABLE~\ref{tab:Quantitative evaluations of state-of-the-art SR methods}. Since the trained model for a scale factor of $3\times$ is not provided by  LapSRN  \cite{29LaiWS2017CVPR_LapSRN}, we generate the corresponding results via downscaling its $4\times$ upscaling results as the way in  \cite{29LaiWS2017CVPR_LapSRN}. While the proposed CMSC\_SR achieves comparable results to state-of-the-art approaches, our final model CMSC significantly outperforms all exiting methods on all datasets for all upscaling factors, in terms of PSNR, SSIM and IFC.  Compared to  MemNet \cite{22TaiY2017ICCV_MemNet} which obtains the highest performances among the prior methods, our proposed CMSC achieves the improvements of 0.12dB, 0.11dB and 0.12dB respectively for three upscaling factors ($2\times$, $3\times$, $4\times$) on the average PSNRs of four datasets. Especially, on the very challenging dataset Urban100, the proposed CMSC outperforms the state-of-the-art method (MemNet \cite{22TaiY2017ICCV_MemNet}) by the PSNR gains of 0.16dB, 0.13dB and 0.14 dB on scale factors of $2\times$, $3\times$ and $4\times$ respectively. In addition, objective image quality assessment values in terms of SSIM and IFC scores further validate the superiority of the proposed method.

\begin{figure*}
\vspace{-0.4cm}
\centering
\captionsetup{belowskip=-14pt,aboveskip=-0.005pt,width=0.99\linewidth}
\includegraphics[width=0.99\linewidth]{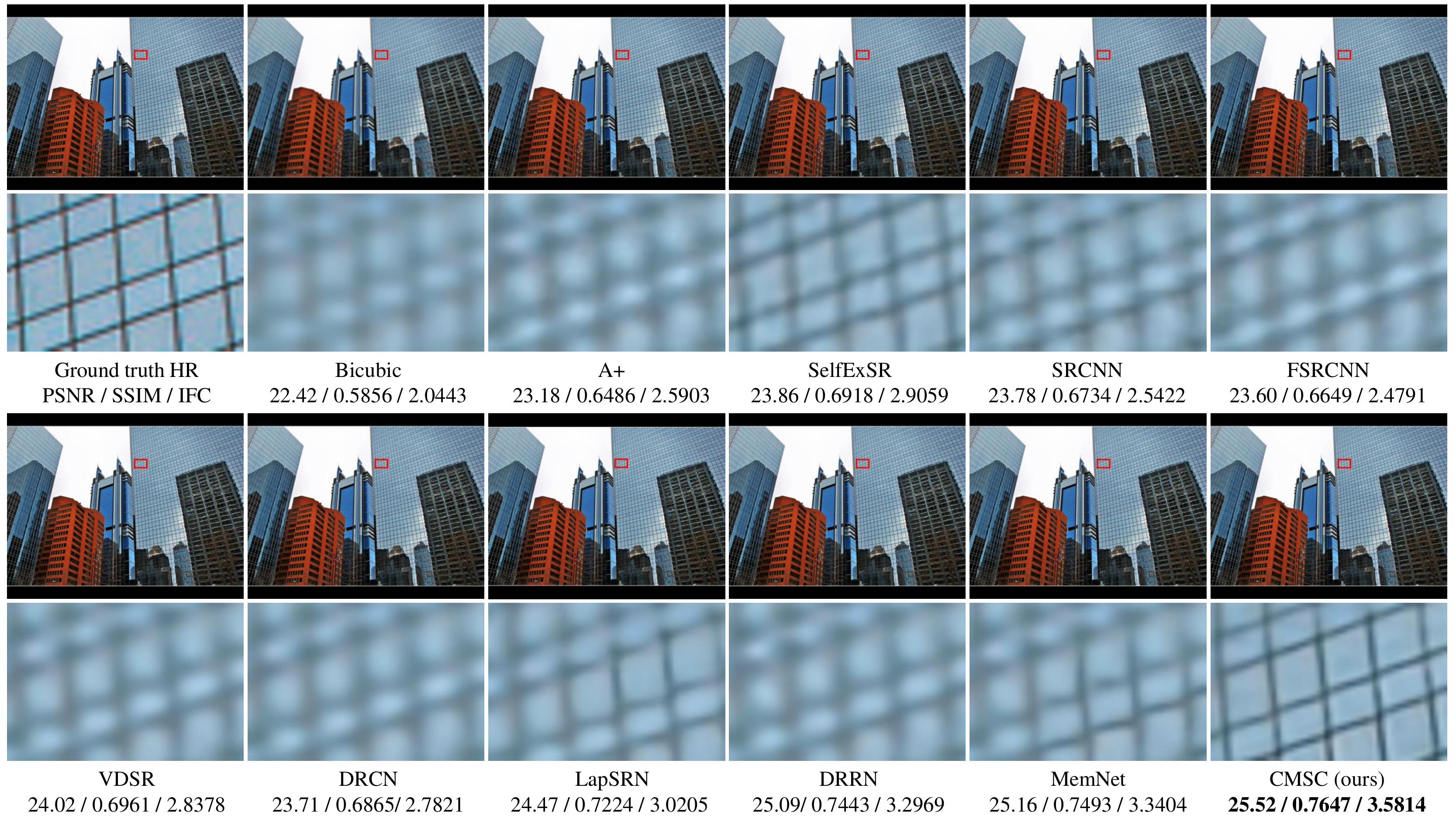}
\caption{Visual evaluation for a scale factor of $4\times$ on the ``img099'' image  from Urban100. The CMSC reconstructs the shaper grid lines which are very blurry in other results.}
\label{fig: x4 img099}
\end{figure*}

The visual comparisons of different methods are shown in Fig.\ref{fig: x3 zebra}, Fig.\ref{fig: x4 8023}, Fig.\ref{fig: x4 img053} and Fig.\ref{fig: x4 img099}. Our proposed CMSC accurately and clearly reconstructs the texture pattern, the grid pattern and the lines. It is observed that the severe distortions and the artifacts are contained in the results generated by the prior methods, such as the marked zebra stripes and texture regions of the wing in Fig.\ref{fig: x3 zebra} and Fig.\ref{fig: x4 8023}. In contrast, our method avoids the distortions and suppresses the artifacts via the cascaded features reconstruction, the residual-features learning and the multi-scale information fusion. In addition, in Fig.\ref{fig: x4 img053} and Fig.\ref{fig: x4 img099}, only our method is able to reconstruct finer edges and clearer grids while other methods generate very blurry results.

We also adopt the public source codes of state-of-the-art methods to measure the execution time. Since the testing codes of  SRCNN \cite{11Dong2016TPAMI}  and FSRCNN \cite{46Dong2016ECCVFSRCNN} are implemented on the CPU, we then rebuild both models as well as the VDSR \cite{12KimJ2016CVPR_VDSR} model in PyTorch with the same network parameters for evaluating the runtime on GPU. Fig.\ref{fig:PSNR performance versus runtime} shows the PSNR  performance  versus  execution time in the testing phase on the  Set5  dataset  for a scale factor of $2\times$. We can see that our proposed CMSC outperforms all mentioned methods with relatively less execution time. Our source code will be released to the  public later.

\section{Conclusion}
In this paper, we propose a deep cascaded multi-scale cross network (CMSC) for modeling the super-resolution reconstruction process, where a sequence of subnetworks is cascaded to gradually refine high resolution features with cascaded-supervision in a coarse-to-fine manner. In each cascaded subnetwork, multiple multi-scale cross (MSC) modules are stacked not only to fuse complementary information under different receptive fields but also to improve information flow across the layers. Besides, to make full use of relative information between high-resolution  and low-resolution features, residual-features learning is introduced to the cascaded subnetworks for further boosting reconstruction performance. Comprehensive evaluations on benchmark datasets demonstrate that our CMSC network outperforms  state-of-the-art super-resolution methods in terms of quantitative and qualitative evaluations  with relatively less execution time.

Since the subnetworks in CMSC at all stages have the same structure and the same aim, it is possible for our model to share the network parameters across the cascaded stages. In future work, we will explore a suitable strategy to share the parameters across as well as within the cascaded stages, and thus to control the number of model parameters without a decrease in performance. On the other hand, we will extend our CMSC model to other image restoration and heterogeneous image transformation fields.

{
\bibliographystyle{IEEEtran}
\bibliography{CMSC}

\begin{thebibliography}{10}
\providecommand{\url}[1]{#1}
\csname url@samestyle\endcsname
\providecommand{\newblock}{\relax}
\providecommand{\bibinfo}[2]{#2}
\providecommand{\BIBentrySTDinterwordspacing}{\spaceskip=0pt\relax}
\providecommand{\BIBentryALTinterwordstretchfactor}{4}
\providecommand{\BIBentryALTinterwordspacing}{\spaceskip=\fontdimen2\font plus
\BIBentryALTinterwordstretchfactor\fontdimen3\font minus
  \fontdimen4\font\relax}
\providecommand{\BIBforeignlanguage}[2]{{%
\expandafter\ifx\csname l@#1\endcsname\relax
\typeout{** WARNING: IEEEtran.bst: No hyphenation pattern has been}%
\typeout{** loaded for the language `#1'. Using the pattern for}%
\typeout{** the default language instead.}%
\else
\language=\csname l@#1\endcsname
\fi
#2}}
\providecommand{\BIBdecl}{\relax}
\BIBdecl

\bibitem{1Freeman2000IJCV}
W.~T. Freeman, E.~C. Pasztor, and O.~T. Carmichael, ``Learning low-level
  vision,'' \emph{Int. J. Comput. Vis.}, vol.~40, no.~1, pp. 25--47, Oct. 2000.

\bibitem{2Polatkan2015TPAMI}
G.~Polatkan, M.~Zhou, L.~Carin, and D.~Blei, ``A {Bayesian} non-parametric
  approach to image super-resolution,'' \emph{IEEE Trans. Pattern Anal. Mach.
  Intell.}, vol.~37, no.~2, pp. 346--358, Feb. 2015.

\bibitem{3Chang2004CVPR}
H.~Chang, D.-Y. Yeung, and Y.~Xiong, ``Super-resolution through neighbor
  embedding,'' in \emph{Proc. IEEE Conf. Comput. Vis. Pattern Recognit.
  (CVPR)}, Jun./Jul. 2004, pp. 275--282.

\bibitem{4Gao2012TIP}
X.~Gao, K.~Zhang, D.~Tao, and X.~Li, ``Image super-resolution with sparse
  neighbor embedding,'' \emph{IEEE Trans. Image Process.}, vol.~21, no.~7, pp.
  3194--3205, Jul. 2012.

\bibitem{5Yang2010TIP}
J.~Yang, J.~Wright, T.~S. Huang, and Y.~Ma, ``Image super-resolution via sparse
  representation,'' \emph{IEEE Trans. Image Process.}, vol.~19, no.~11, pp.
  2861--2873, Nov. 2010.

\bibitem{6He2013CVPR}
L.~He, H.~Qi, and R.~Zaretzki, ``Beta process joint dictionary learning for
  coupled feature spaces with application to single image super-resolution,''
  in \emph{Proc. IEEE Conf. Comput. Vis. Pattern Recognit. (CVPR)}, Jun. 2013,
  pp. 345--352.

\bibitem{7Timofte2014ACCV}
R.~Timofte, V.~D. Smet, and L.~V. Gool, ``A+: adjusted anchored neighborhood
  regression for fast super-resolution,'' in \emph{Proc. 12th Asian Conf.
  Comput. Vis. (ACCV)}, Nov. 2014, pp. 111--126.

\bibitem{8Hu2016TIP}
Y.~Hu, N.~Wang, D.~Tao, X.~Gao, and X.~Li, ``{SERF}: a simple, effective,
  robust, and fast image super-resolver from cascaded linear regression,''
  \emph{IEEE Trans. Image Process.}, vol.~25, no.~9, pp. 4091--4102, Sep. 2016.

\bibitem{9Wang2016TIP}
H.~Wang, X.~Gao, K.~Zhang, and J.~Li, ``Single image super-resolution using
  active-sampling {Gaussian} process regression,'' \emph{IEEE Trans. Image
  Process.}, vol.~25, no.~2, pp. 935--948, Feb. 2016.

\bibitem{10Schulter2015CVPR}
S.~Schulter, C.~Leistner, and H.~Bischof, ``Fast and accurate image upscaling
  with super-resolution forests,'' in \emph{Proc. IEEE Conf. Comput. Vis.
  Pattern Recognit. (CVPR)}, Jun. 2015, pp. 3791--3799.

\bibitem{11Dong2016TPAMI}
C.~Dong, C.~C. Loy, K.~He, and X.~Tang, ``Image super-resolution using deep
  convolutional networks,'' \emph{IEEE Trans. Pattern Anal. Mach. Intell.},
  vol.~38, no.~2, pp. 295--307, Feb. 2016.

\bibitem{12KimJ2016CVPR_VDSR}
J.~Kim, J.~K. Lee, and K.~M. Lee, ``Accurate image super-resolution using very
  deep convolutional networks,'' in \emph{Proc. IEEE Conf. Comput. Vis. Pattern
  Recognit. (CVPR)}, Jun. 2016, pp. 1646--1654.

\bibitem{13KimJ2016CVPR_DRCN}
{J. Kim, and J. K. Lee, and K. M. Lee}, ``Deeply-recursive convolutional
  network for image super-resolution,'' in \emph{Proc. IEEE Conf. Comput. Vis.
  Pattern Recognit. (CVPR)}, Jun. 2016, pp. 1637--1645.

\bibitem{14TaiY2017CVPR_DRRN}
Y.~Tai, J.~Yang, and X.~Liu, ``Image super-resolution via deep recursive
  residual network,'' in \emph{Proc. IEEE Conf. Comput. Vis. Pattern Recognit.
  (CVPR)}, Jul. 2017, pp. 3147--3155.

\bibitem{15HeK2016ECCV_identity}
K.~He, X.~Zhang, S.~Ren, and J.~Sun, ``Identity mappings in deep residual
  networks,'' in \emph{Proc. Eur. Conf. Comput. Vis. (ECCV)}, Sep. 2016, pp.
  630--645.

\bibitem{16LedigC2017CVPR_SRResNet}
C.~Ledig, L.~Theis, F.~Huszar, J.~Caballero, A.~P. Aitken, A.~Tejani, J.~Totz,
  Z.~Wang, and W.~Shi, ``Photo-realistic single image super-resolution using a
  generative adversarial network,'' in \emph{Proc. IEEE Conf. Comput. Vis.
  Pattern Recognit. (CVPR)}, Jul. 2017, pp. 105--114.

\bibitem{17Lim2017CVPRW_EDSR}
B.~Lim, S.~Son, H.~Kim, S.~Nah, and K.~M. Lee, ``Enhanced deep residual
  networks for single image super-resolution,'' in \emph{Proc. IEEE Conf.
  Comput. Vis. Pattern Recognit. Workshops (CVPRW)}, Jul. 2017, pp. 136--144.

\bibitem{18Timofte2017CVPRW_NTIRE}
R.~Timofte, E.~Agustsson, L.~V. Gool, M.-H. Yang, L.~Zhang, B.~Lim, and et~al.,
  ``{NTIRE} 2017 challenge on single image super-resolution: methods and
  results,'' in \emph{Proc. IEEE Conf. Comput. Vis. Pattern Recognit. Workshops
  (CVPRW)}, Jul. 2017, pp. 1110--1121.

\bibitem{19Mao2016NIPS_RED}
X.-J. Mao, C.~Shen, and Y.-B. Yang, ``Image restoration using very deep
  convolutional encoder-decoder networks with symmetric skip connections,'' in
  \emph{Proc. Adv. Neural Inf. Process. Syst. (NIPS)}, Dec. 2016, pp.
  2802--2810.

\bibitem{20Huang2017CVPR_DenseNet}
G.~Huang, Z.~Liu, K.~Q. Weinberger, and L.~van~der Maaten, ``Densely connected
  convolutional networks,'' in \emph{Proc. IEEE Conf. Comput. Vis. Pattern
  Recognit. (CVPR)}, Jul. 2017, pp. 4700--4708.

\bibitem{21Tong2017ICCV_DenseSR}
T.~Tong, G.~Li, X.~Liu, and Q.~Gao, ``Image super-resolution using dense skip
  connections,'' in \emph{Proc. IEEE Int. Conf. Comput. Vis. (ICCV)}, Oct.
  2017, pp. 4799--4807.

\bibitem{22TaiY2017ICCV_MemNet}
Y.~Tai, J.~Yang, X.~Liu, and C.~Xu, ``{MemNet:} a persistent memory network for
  image restoration,'' in \emph{Proc. IEEE Int. Conf. Comput. Vis. (ICCV)},
  Oct. 2017, pp. 4549--4557.

\bibitem{23WangY2016arXiv}
Y.~Wang, L.~Wang, H.~Wang, and P.~Li, ``End-to-end image super-resolution via
  deep and shallow convolutional networks,'' \emph{arXiv: 1607.07680}, Jul.
  2016.

\bibitem{24Tang2018Neurocomputing}
Z.~Tang, L.~Luo, H.~Peng, and S.~Li, ``A joint residual network with paired
  {ReLUs} activation for image super-resolution,'' \emph{Neurocomputing}, vol.
  273, pp. 37--46, Jan. 2018.

\bibitem{25Yamanaka2017ICONIP}
J.~Yamanaka, S.~Kuwashima, and T.~Kurita, ``Fast and accurate image super
  resolution by deep {CNN} with skip connection and network in network,'' in
  \emph{Int. Conf. Neural Inf. Process. (ICONIP)}, Nov. 2017, pp. 217--225.

\bibitem{26Ren2017CVPRW}
H.~Ren, M.~El-Khamy, and J.~Lee, ``Image super resolution based on fusing
  multiple convolution neural networks,'' in \emph{IEEE Conf. Comput. Vis.
  Pattern Recognit. Workshops (CVPRW)}, Jul. 2017, pp. 54--61.

\bibitem{27Wang2015ICCV}
Z.~Wang, D.~Liu, J.~Yang, W.~Han, and T.~S. Huang, ``Deep networks for image
  super-resolution with sparse prior,'' in \emph{Proc. IEEE Int. Conf. Comput.
  Vis. (ICCV)}, Dec. 2015, pp. 370--378.

\bibitem{28Cui2014ICCV}
Z.~Cui, H.~Chang, S.~Shan, B.~Zhong, and X.~Chen, ``Deep network cascade for
  image super-resolution,'' in \emph{Proc. Eur. Conf. Comput. Vis. (ECCV)},
  Sep. 2014, pp. 49--64.

\bibitem{29LaiWS2017CVPR_LapSRN}
W.-S. Lai, J.-B. Huang, N.~Ahuja, and M.-H. Yang, ``Deep {Laplacian} pyramid
  networks for fast and accurate super-resolution,'' in \emph{Proc. IEEE Conf.
  Comput. Vis. Pattern Recognit. (CVPR)}, Jul. 2017, pp. 624--632.

\bibitem{46Dong2016ECCVFSRCNN}
C.~Dong, C.~C. Loy, and X.~Tang, ``Accelerating the super-resolution
  convolutional neural network,'' in \emph{Proc. Eur. Conf. Comput. Vis.
  (ECCV)}, Oct. 2016, pp. 391--407.

\bibitem{30Zagoruyko2017arXivDiracNet}
S.~Zagoruyko and N.~Komodakis, ``{DiracNets:} training very deep neural
  networks without skip-connections,'' \emph{arXiv: 1706.00388}, Jun. 2017.

\bibitem{31Zhao2017arXivMR}
L.~Zhao, J.~Wang, X.~Li, Z.~Tu, and W.~Zen, ``Deep convolutional neural
  networks with merge-and-run mappings,'' \emph{arXiv: 1611.07718}, Jul. 2017.

\bibitem{32Srivastava2015NIPS}
R.~K. Srivastava, K.~Greff, and J.~Schmidhuber, ``Training very deep
  networks,'' in \emph{Proc. Adv. Neural Inf. Process. Syst. (NIPS)}, Dec.
  2015, pp. 2377--2385.

\bibitem{33He2016CVPRResNet}
K.~He, X.~Zhang, S.~Ren, and J.~Sun, ``Deep residual learning for image
  recognition,'' in \emph{IEEE Conf. Comput. Vis. Pattern Recognit. (CVPR)},
  Jun./Jul. 2016, pp. 770--778.

\bibitem{34Szegedy2015CVPR}
C.~Szegedy, W.~Liu, Y.~Jia, P.~Sermanet, S.~Reed, D.~Anguelov, D.~Erhan,
  V.~Vanhoucke, and A.~Rabinovich, ``Going deeper with convolutions,'' in
  \emph{IEEE Conf. Comput. Vis. Pattern Recognit. (CVPR)}, Jun. 2015, pp. 1--9.

\bibitem{35Szegedy2016arXiv}
C.~Szegedy, S.~Ioffe, and V.~Vanhoucke, ``Inception-v4, inception-resnet and
  the impact of residual connections on learning,'' \emph{arXiv: 1602.07261},
  Aug. 2016.

\bibitem{36Szegedy2016CVPR}
C.~Szegedy, V.~Vanhoucke, S.~Ioffe, J.~Shlens, and Z.~Wojna, ``Rethinking the
  inception architecture for computer vision,'' in \emph{IEEE Conf. Comput.
  Vis. Pattern Recognit. (CVPR)}, Jun./Jul. 2016, pp. 2818--2826.

\bibitem{37Timofte2016CVPRSevenWays}
R.~Timofte, R.~Rothe, and L.~V. Gool, ``Seven ways to improve example-based
  single image super resolution,'' in \emph{IEEE Conf. Comput. Vis. Pattern
  Recognit. (CVPR)}, Jun./Jul. 2016, pp. 1865--1873.

\bibitem{38Timofte2015ICMLBN}
S.~Ioffe and C.~Szegedy, ``Batch normalization: accelerating deep network
  training by reducing internal covariate shift,'' in \emph{Int. Conf. Mach.
  Learn. (ICML)}, Jul. 2015, pp. 448--456.

\bibitem{39Maas2013ICMLLeakReLU}
A.~L. Maas, A.~Y. Hannun, and A.~Y. Ng, ``Rectifier nonlinearities improve
  neural network acoustic models,'' in \emph{Int. Conf. Mach. Learn. (ICML)},
  Jun. 2013.

\bibitem{40Bevilacqua2012BMVC}
M.~Bevilacqua, A.~Roumy, C.~Guillemot, and M.-L.~A. Morel, ``Low-complexity
  single-image super-resolution based on nonnegative neighbor embedding,'' in
  \emph{Proc. 23rd British Mach. Vis. Conf. (BMVC)}, Sep. 2012, pp.
  135.1--135.10.

\bibitem{41Zeyde2010ICCS}
R.~Zeyde, M.~Elad, and M.~Protter, ``On single image scale-up using
  sparse-representations,'' in \emph{Proc. 7th Int. Conf. Curves Surfaces},
  Jun. 2010, pp. 711--730.

\bibitem{42ArbelaezP2011TPAMI}
P.~Arbel{\'{a}}ez, M.~Maire, C.~Fowlkes, and J.~Malik, ``Contour detection and
  hierarchical image segmentation,'' \emph{IEEE Trans. Pattern Anal. Mach.
  Intell.}, vol.~33, no.~5, pp. 898--916, May. 2011.

\bibitem{43Huang2015CVPRSelfExp}
J.-B. Huang, A.~Singh, and N.~Ahuja, ``Single image super-resolution from
  transformed self-exemplars,'' in \emph{IEEE Conf. Comput. Vis. Pattern
  Recognit. (CVPR)}, Jun. 2015, pp. 5197--5206.

\bibitem{47Wang2004TIPSSIM}
Z.~Wang, A.~C. Bovik, H.~R. Sheikh, and E.~P. Simoncelli, ``Image quality
  assessment: from error visibility to structural similarity,'' \emph{IEEE
  Trans. Image Process.}, vol.~13, no.~4, pp. 600--612, Apr. 2004.

\bibitem{44Sheikh2005TIPIFC}
H.~R. Sheikh, A.~C. Bovik, and G.~de~Veciana, ``An information fidelity
  criterion for image quality assessment using natural scene statistics,''
  \emph{IEEE Trans. Image Process.}, vol.~14, no.~12, pp. 2117--2128, Dec.
  2005.

\bibitem{45He2015CVPRInitial}
K.~He, X.~Zhang, S.~Ren, and J.~Sun, ``Delving deep into rectifiers: surpassing
  human-level performance on imagenet classification,'' in \emph{in Proc. IEEE
  Int. Conf. Comput. Vis. (ICCV)}, Dec. 2015, pp. 1026--1034.

\end{thebibliography}
}

\end{document}